\documentclass[journal,onecolumn]{IEEEtran}

\usepackage{lineno}
\usepackage{tabularx}
\usepackage{float}
\usepackage{bm}

\usepackage{cite}

\ifCLASSINFOpdf
   \usepackage[pdftex]{graphicx}
\else
\fi
\usepackage{amsmath}
\usepackage{tabularx}

\usepackage{algorithm}
\usepackage{algpseudocode}

\usepackage{stfloats}
\begin{document}
%

\title{UGSim: Autonomous Buoyancy-Driven Underwater Glider Simulator with LQR Control Strategy and Recursive Guidance System}

%
%
%
\author{Zhizun~Xu, \thanks{Zhizun Xu is with School of Engineering, Newcastle University, Newcastle upon Tyne UK, and with School of Naval Architecture and Maritime, Guangdong Ocean University, China, Zhizun.Xu@newcastle.ac.uk}
Yang~Song,\thanks{Yang Song is with School of Engineering, Newcastle University, Newcastle upon Tyne UK}
Jiabao~Zhu,\thanks{Jiabao Zhu is with School of Naval Architecture and Maritime, Guangdong Ocean University, China}
and 
Weichao~Shi\thanks{Weichao Shi(Corresponding author) is with School of Engineering, Newcastle University, Newcastle upon Tyne UK, Weichao.Shi@newcastle.ac.uk}
}

\maketitle

\begin{abstract}

This paper presents the UGSim, a simulator for buoyancy-driven gliders, with a LQR control strategy, and a recursive guidance system. Building on the top of the DAVE and the UUVsim, it is designed to address unique challenges that come from the complex hydrodynamic and hydrostatic impacts on buoyancy-driven gliders, which conventional robotics simulators can't deal with. Since distinguishing features of the class of vehicles, general controllers and guidance systems developed for underwater robotics are infeasible. The simulator is provided to accelerate the development and the evaluation of algorithms that would otherwise require expensive and time-consuming operations at sea. It consists of a basic kinetic module, a LQR control module and a recursive guidance module, which allows the user to concentrate on the single problem rather than the whole robotics system and the software infrastructure. We demonstrate the usage of the simulator through an example, loading the configuration of the buoyancy-driven glider named Petrel-II, presenting its dynamics simulation, performances of the control strategy and the guidance system.

\end{abstract}


\IEEEpeerreviewmaketitle

\section{Introduction}
\label{sec:intro}

In autonomous buoyancy-driven underwater gliders, electric or thermal energy is converted to pressure-volume work to change the vehicle volume to cycle vertically in the ocean. They use lift on wings to project the vertical velocity into forward motion. These gliders are steered by changing the centre of gravity with respect to the centre of the buoyancy, thus controlling both the pitch and the roll. Because of those special designs, there are four distinguishing characteristics for the class of vehicles are: 1)the use of buoyancy, 2)a sawtooth operating pattern, 3)long duration, 4)relatively slow operating speeds\cite{davis2002autonomous}. These features make the vehicles be ideal tools for sampling vertical data of the interior ocean. They have huge amount of potential in oceanographic sensing missions\cite{graver2003underwater}, military reconnaissance operations, and marine exploration applications\cite{arshad2012buoyancy}.

Due to above merits, main oceanographic research institutions have put their efforts and resources to accelerate the development of this cutting-edging technology. Advent of material and microelectronic technical revolutions, the body of the vehicle are able to endure extreme conditions introduced by the uncertain ocean environment, the electronics are supposed to be reliable and able to drive the buoyancy pump, change the position of the battery, and collect data from onboard sensors. Some of them have been put into commercial productions\cite{webb2001slocum}\cite{sherman2001autonomous}\cite{xue2018coordinate}. However, correspondent researchers lack a high quality simulator for helping software developments.

In general robotics field, a high quality numerical simulator, which estimates robotic motions, interacts with the virtual environment and communicates with sensors in a realistic fashion, enables researchers and developers to develop the guidance, the navigation and the control software without the need for the advanced hardware to test their innovations\cite{henriksen2016uw}. This is expected to lower the cost, making it faster and easier to test new software and making it possible to study how robots interact with the environment without the risk of breaking equipment. Compared with ground robotics and drones, the scenarios for underwater glider operation have three Rs\cite{zhang2022dave}: remote, risky, and recalcitrant. The gliders are supposed to be evaluated in deep ocean, receiving commands remotely via satellite signals. These field trials usually require oceanographic support vessels and large logistical requirements for the deployment and the recovery. That makes the physical testing particularly costly. Furthermore, for gliders are designed for long-endurance data collecting missions, they are expected to travel along the sawtooth pattern in ocean and maintain feeding back measurement data for weeks or months, even years. Such long term physical tests are conducted to guarantee their proper functioning. However, these operations involve significant risks because undiscovered software bugs might cause the losing of the robotic platform. Testing gliders in various environmental conditions is another challenge. The interior ocean environments are recalcitrant in the sense that the software in gliders must be designed for a wide range of oceanographic conditions, but it is impossible for operators to regulate environmental conditions where they are launched. Hence, as for autonomous buoyancy-driven underwater gliders, simulator is especially valuable as a complement to perfect the software prior to field trials. 

Unfortunately, existing underwater vehicle simulators are not designed to deal with the unique challenges coming with distinctive features of autonomous buoyancy-driven underwater gliders, including the special propulsion and steering mechanism, and complex hydrodynamic and hydrostatic effects. General physics engines intend to tackle of rigid-body dynamic simulations, while hydrodynamic and hydrostatic forces and moments must also be taken into account. The computation of the interaction of a fluid with submerged body is still infeasible for an off-the-shelf computer for complex geometries like underwater vehicles in real-time\cite{cieslak2019stonefish}. Conventional underwater robotics simulators \cite{manhaes2016uuv}\cite{cieslak2019stonefish}\cite{zhang2022dave}\cite{henriksen2016uw} mainly rely on the Fossen's model\cite{fossen2011handbook} to estimate the robotic motion impacted by the fluid, which has been proved successfully in ordinary underwater robotics. However, due to their unique features, the Fossen's model is hard to describe the motion of autonomous buoyancy-driven underwater gliders. 

Attempting to bridge the gap and offer a high-quality tool in the software development, a novel simulator of autonomous buoyancy-driven underwater gliders, named UGSim, has been proposed. The simulator is built on top of the UUV simulator and the DAVE project, taking advantages of their merits, including the multi-robots collaboration, the stratified ocean currents simulation, the large scale environment generation, and the dynamic bathymetry spawning. The simulator employs the Gazebo\cite{koenig2004design} as a physic engine, tackling rigid-body dynamics and the collision detection. It also works as a 3D rendering pipeline, visualising subsea environments and offshore engineering scenarios. The simulator utilises the Gazebo model plugin API(Application Programming Interface) to impose hydrodynamic and hydrostatic effects on gliders. As mentioned before, behaviour patterns of buoyancy-driven gliders are huge different with ordinary ROVs and AUVs. That means most controllers and guidance systems for underwater vehicles are unsuitable. In order to reduce software development workload, and avoid reinventing wheels, the paper provides novel control and guidance modules. Integrating the dead-reckoning module offered by Zhang \cite{zhang2022dave}, the basic waypoints-tracking function can be achieved.

Hence, the contributions of the paper are:
 \begin{itemize}
     \item A novel motion simulation of buoyancy-driven gliders with environmental interactions, including collisions, and external disturbances(such as current); 
     \item A LQR control strategy of the buoyancy-driven glider in consecutive work cycles;
     \item A novel recursive guidance system of the buoyancy-driven glider, which is designed for its features: the long-term operation, the low manoeuvrability, and the low accuracy underwater positioning.
 \end{itemize}

\section{Literature Review}

The underwater simulator has been developed for decades. The UWsim is one of the first and most commonly known underwater simulator\cite{prats2012open}, it offers the simulation platform meeting needs of the underwater inspection and intervention missions with one or more robots. The simulator utilised OpenSceneGraph(OSG) and osgOcean libraries to render realistic underwater environments. The former one is an open source 3D graphics application programming interface used by application developers in fields such as the visual simulation. The latter one is another open source project that implements the realistic underwater rendering using the OSG. The UWSim uses the above mentioned libraries and adds the further functionality for easily adding underwater robotics to the scene, and sensors simulations. It has interfaces with external control programs through the Robot Operating System(ROS). However, it has been noted that while it is advantageous in visualization and sensors' simulation, it is lacking in dynamic simulation\cite{cieslak2019stonefish}. The kinematic and kinetic module based on Fossen's models are coded in MATLAB. The output variables(position and attitude of vehicles) from MATLAB codes are broadcasted through the ROS network and captured by the UWSim core for updating the visualisation.

In the mobile robotics community, the Gazebo is a very popular simulator, which is designed for general-purpose open-source robotics. The Gazebo was used in a humanoid robotic project called DexROV\cite{birk2018dexterous}, coupled with a set of plugins. The Gazebo Classic utilises the Open Dynamic Engine(ODE) as its default physics engine with three alternative physics engines: Bullet\cite{coumans2015bullet}, Simbody\cite{sherman2011simbody} and DART\cite{lee2018dart}. The 3D rendering pipeline in the Gazebo Classic is implemented by the Object-Oriented Graphics Rendering Engine(OGRE), which is a scene-oriented, real-time, open-source, 3D rendering engine. The controversy over the Gazebo is that it lacks native supports for hydrodynamics and its poor rendering quality which does not allow for the convincing simulation of the underwater vision. But the Gazebo prioritizes physics over rendering, offering the higher physical confidence. 

Hence, in order to overcome inferior dynamic performances in the UWSim, freefloating-gazebo\cite{kermorgant2014dynamic} was presented as an example of a bridge between the Gazebo and the UWSim for simulations of underwater vehicles. The package includes plugins for the Gazebo to allow the generation of hydrodynamic and hydrostatic forces to be applied on underwater vehicles. The estimated pose of the vehicle will be transmitted to the UWSim via the ROS, where the better rendering quality can be achieved. 

A similar implementation involves the extension of the Gazebo to the Robot Construction Kit(ROCK), known as the ROCK-Gazebo\cite{watanabe2015rock}. In this case, the ROCK visualisation tool was extended with \textit{OpenSceneGraph}\cite{burns2004open} for rendering the underwater environment while the physic simulation runs in the Gazebo. At this moment, it does not yet support the simulation of multiple underwater vehicles. 

The latest gazebo-based underwater simulator is the DAVE project\cite{zhang2022dave}, which is developed on the top of the UUV simulator\cite{manhaes2016uuv}. Zhang et al. provided a simulation framework for the multiple-types marine craft, including surface ships, ROVs, AUVs, and gliders. The highlights of the simulator includes: 1) the definition of stratified ocean currents by directions and depths either constant or periodic in time; 2) the dynamic bathymetry spawning, which means that grid tiles produced from the large, high-resolution bathymetry height map data are dynamically spawned and unloaded. It will increase the efficiency of the memory usage and contribute to accelerate the large scale world simulation.

Prior to the DAVE project, the UUV simulator\cite{manhaes2016uuv} is well-known for offering a set of newly implemented plugins that model underwater hydrostatic and hydrodynamic effects, thrusters, sensors, and external disturbances. It was designed originally to meet the need of the EU-Funded project SwarMs, and to support the development of new missions strategies and high-level algorithms for cooperative behaviours of underwater vehicles.

Additionally, the WHOI Deep Submergence Lab \cite{dssim} released common tools and a multi-sensors package, named \textit{ds\_sim}. The package composes the Gazebo binary plugins for a number of useful sensors and some utilities(such as DVL, depth sensor, and USBL SMS device, and GPS). In this case, other Gazebo-based underwater vehicle simulators are able to make use of the multi-sensors plugin directly.

Because the Gazebo does not allow for a camera image with a degraded visibility which frequently happens in the subsea and results from the back refraction, Suzuki and Kawabta\cite{suzuki2020development} reported two plugins named as \textit{FluidDynamicsPlugin} and \textit{ImageNoisePlugin} on a generic platform named Choreonoid\cite{nakaoka2012choreonoid}, simulating the fluid dynamics, the buoyancy, the fluid resistance and the visual distortion caused by the camera lens. 

Boosting the 3D rendering quality, there are graphics- and game-driven simulators which trade off the physical fidelity with the advanced rendering. It is a necessary balance for all simulators with finite computing resource. Sonefish\cite{cieslak2019stonefish} directly uses the OpenGL to provide the lightweight, high performance rendering pipeline. Its unique feature is being able to simulate complete dynamics or hydrodynamics of underwater vehicle-manipulator systems with contact and force sensing. Its dynamic simulation framework is a set of classes that wrap around the physics library(Bullet Physics) adding an abstraction layer which greatly simplifies creation of simulation scenarios and adds features related to robotics, especially marine robotics, e.g., the buoyancy and hydrodynamics. 

Game engines known for the photo-realistic rendering have also been chosen as tools for the visualization of the underwater simulation. The Unreal Engine\cite{unreal} has been used in vision-based applications. The Unity3D\cite{unity} has been used in simulators such the URSim\cite{prats2012open}. Hydrodynamic forces achieved by using C\# scripts that make used of Unity's APIs to emulate forces on the object's mesh to account for an external physical influence on the vehicle, but limited sensors supports. The MORSE\cite{echeverria2011modular} rendering with the Blender Game Engine, a generic simulator for academic robotics. The UW MORSE\cite{henriksen2016uw}, a underwater vehicle simulator, is developed based on the MORSE, translating damping forces, the buoyancy, and thruster forces into vectors relative to world frame and apply them to the rigid body of the vehicle in the Bullet physic engine. It has multiple sensors support, with interfaces for the MOOS—IvP and the ROS.

Apart from visual tools, there is also the Marine System Simulator\cite{mss} which is a MATLAB/Simulink package. However, lacking of the visualisation, and no direct interfaces with the ROS make it unpopular in the community.

However, existing robotic simulators cannot estimate the motion of buoyancy-driven gliders. The paper proposes a simulator specified for the class of the vehicles with a control strategy and a guidance system. As the 3D rendering is not emphasised in glider simulations, the simulator is developed based on the DAVE project, the UUV simulator, and \textit{ds\_sim}, by adding the control module, the guidance module, and the kinetic module.

\section{Simulator Framework}

The Gazebo is considered as an accurate framework for simulation and visualization for robotics mechanisms, can be extended to new dynamics, sensors and world models through modular plugins\cite{koenig2004design}. It provides developers APIs(Application Programming Interfaces) for user-define plugins of models, sensors, and visuals. These plugins have access to the simulation objects and data, can transmit information via topics by using Protocol Buffer3 messages and apply torques and forces to objects in the scenario\cite{manhaes2016uuv}. In order to make use of customised plugins, the robot description files in SDF format\cite{koenig2004design}(an XML format designed for Gazebo) should declare the involved plugins. In order to integrate with ROS, the robot descriptions are represented in URDF or xacro\cite{URDF}. 

The UUV simulator and the DAVE are open-source and Gazebo-based underwater vehicles simulators. They have implemented the fundamental functions, but lack the ability to estimate the manoeuvrability of the buoyancy-driven underwater glider. The proposed simulator is constructed at the top of the UUV simulator and the DAVE by adding a novel kinetic module for the special class of the vehicle. In order to be compatible with the collision detection and rigid body dynamics by the Gazebo build-in physics engine, the accelerations including rotation and translation, computed from the kinetic module, are fed into model status in the Gazebo. That means the equivalent force and torque, caused by hydrodynamics and hydrostatics effects, act on the rigid body. These are counted as initial conditions for rigid dynamic estimations by the build-in physic engine to update the status of the model. 

Since manoeuvre mechanics are significantly different with common ROVs, AUVs, and surface ships. The proposed simulator also provides the basic control module and the guidance module to accelerate the software development. These two modules are written by Python, and wrapped as ROS nodes, which are capable of interaction with the Gazebo server. As for the kinetic module, it is a Gazebo model plugin written in C++, using ROS libraries which allow the plugin to publish model states, and subscribe updates from the ROS node of the control module. In a word, the kinetic module can be regarded as a bridge between the Gazebo server and the ROS master, by using APIs of the Gazebo model and ROS libraries. 

\begin{figure}[H]
    \centering
    \includegraphics[width=1\linewidth]{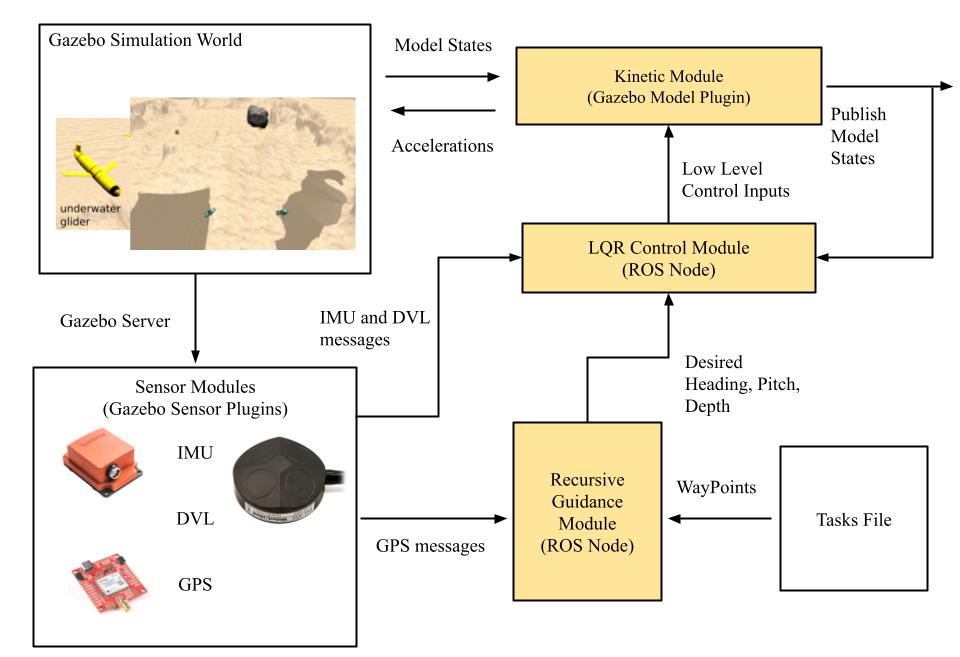}
    \caption{Simulator Framework}
    \label{fig:simulatorFramework}
\end{figure}

In Fig. \ref{fig:simulatorFramework}, the yellow coloured blocks indicate the modules proposed in the paper. The environment models and sensors plugins are provided by the DAVE project. The recursive guidance module is supposed to read waypoints, desired velocities and the target depth from a task file, and produce the desired heading angle and the desired pitch angle, which will be fed to the LQR control module. Note that in sake of the convenience of tests and evaluations, the control module can also receive information directly from the kinetic module. Once receiving sensor messages from sensor plugins, the controller will calculate the LQR control gains, then employ the feedback control law to send low-level control inputs to the kinetic module.

\section{Kinetic Module}

\begin{figure}[H]
    \centering
    \includegraphics[width=1\linewidth]{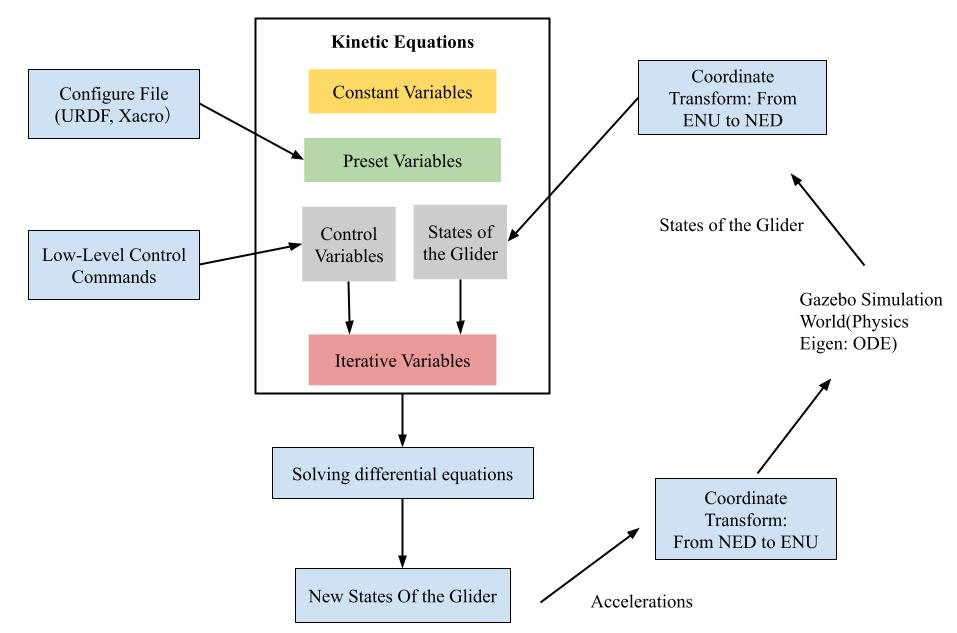}
    \caption{Kinetic Module}
    \label{fig:kineticModule}
\end{figure}

The aim of the kinetic module is to bring hydrodynamic and hydrostatic effects into the rigid body dynamic simulation processed by the Gazebo build-in physics engine. In order to accomplish the goal, the kinetic equations of buoyancy-driven gliders are supposed to be identified firstly. Subsequently, since the kinetic equations are derived in the NED(North-East-Down) frame, rather the ENU(East-North-Up) frame which is applied in Gazebo, the coordinate transformation matrix is required. Finally, the accelerations of linear and angular velocities computed through solving kinetic equations, are fed into model(the glider) status via Gazebo APIs before each rigid-body dynamic simulation iteration. The framework of the kinetic module is shown in Fig.\ref{fig:kineticModule}. The variables in kinetic equations are categorised into five groups: constant variables, preset variables, control variables, states of the glider, and iterative variables.

\subsection{Kinetic Equations}

The kinetic equations were derived in Leonard and Graver's work\cite{leonard2001model}, using Newton law. While the kinetic model is intended to design the SLOCUM glider control algorithm, it is valid and is true by inspections \cite{song2023evaluation}. These kinetic equations are represented below.

\begin{multline}
\begin{bmatrix}
m_s\bm{I} + m_p \bm{I} & -m_s\hat{\bm{r}}_s - m_p \hat{\bm{r}}_p \\
m_s\hat{\bm{r}}_s + m_p\hat{\bm{r}}_p &  (\bm{J}_s + \bm{J}_p) - m_p \hat{\bm{r}}_p \hat{\bm{r}}_p
\end{bmatrix}
\begin{bmatrix}
\dot{\bm{v}}\\
\dot{\bm{\omega}}
\end{bmatrix} =
m_s\begin{bmatrix}
\bm{v}_s \times \bm{\omega}\\
\bm{r}_s\times\bm{v}\times\bm{\omega} + \bm{\omega}\times\bm{r}_s\times\bm{v}
\end{bmatrix} +
\begin{bmatrix}
\bm{0}\\
(\bm{J}_s\bm{\omega})\times\bm{\omega}
\end{bmatrix} \\
+m_p\begin{bmatrix}
\bm{v}_p\times\bm{\omega} + \dot{\bm{r}}_p\times\bm{\omega} \\
\bm{r}_p\times(\bm{v}_p\times\bm{\omega} + \dot{\bm{r}}_p \times \bm{\omega})
\end{bmatrix}
-
\begin{bmatrix}
m_p \ddot{\bm{r}}_p \\
\bm{J}_p\ddot{\zeta}\bm{b}_1 + m_p\bm{r}_p\times\ddot{\bm{r}}_p
\end{bmatrix}
+
\begin{bmatrix}
\bm{0}\\
(\bm{J}_p\bm{\omega}_p)\times\bm{\omega}_p
\end{bmatrix}
+
\begin{bmatrix}
\bm{0}\\
\bm{J}_p(\dot{\bm{\zeta}}\bm{b}_1\times\bm{\omega})
\end{bmatrix}
+
\begin{bmatrix}
\bm{F}_{ext}\\
\bm{T}_{ext}
\end{bmatrix}
\label{eq:kineticsEquations}
\end{multline}

\begin{figure}
    \centering
    \includegraphics[width=1\linewidth]{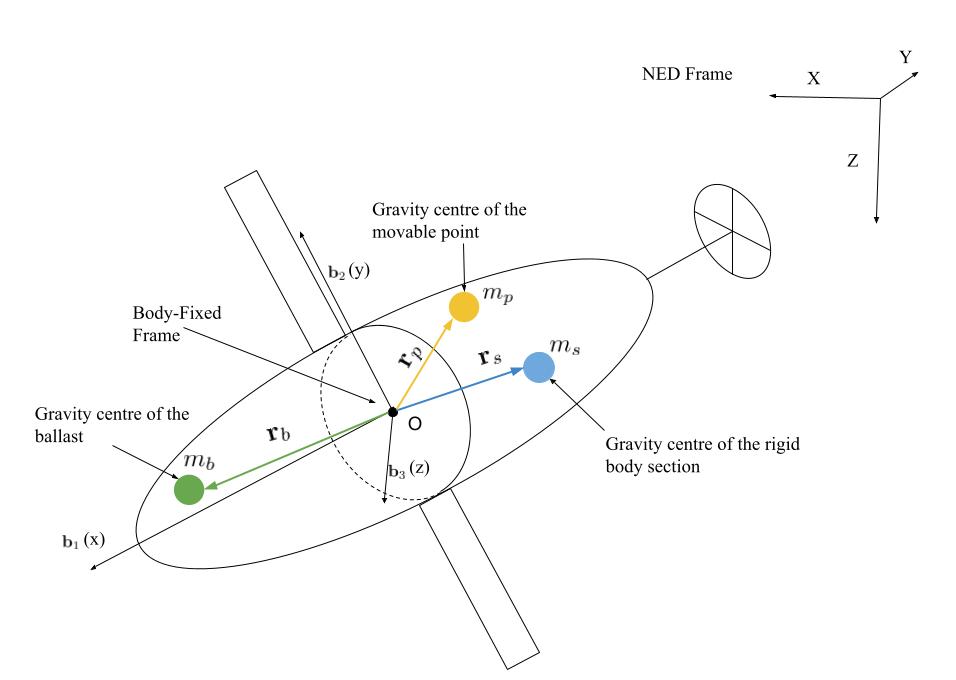}
    \caption{Mass Definitions for Buoyancy-Driven Gliders}
    \label{fig:kineticSketch}
\end{figure}

The definitions of variables are described in Table \ref{tab:abbreviationExplanation}. Mass definitions for buoyancy-driven gliders are shown in Fig.\ref{fig:kineticSketch}

\begin{table}[H]
    \centering
    \caption{Definition of Variables}
    \label{tab:abbreviationExplanation}
    \begin{tabularx}{0.7\textwidth}{c||X}
    \hline
    Name   & Description \\
    \hline
   $\bm{v}$  & linear velocity of the body-fixed frame with respect to the North-East-Down(NED) frame observed in the body-fixed frame \\
   $\bm{\omega}$ & rotation velocity of the body-fixed frame with respect to the NED frame observed in the body-fixed frame \\
   $m_s$ & mass of the rigid body section(excluding the movable point)\\
   $m_p$ & mass of the movable point(the controllable moving part inside the glider body. It usually indicates the battery package.) \\
   $m_b$ & mass of the ballast(It is equivalent to the buoyancy resulted from the variance of volume.)\\
   $\bm{r}_b$ & position of the gravity centre of the ballast($\bm{r}_b=[r_{b1},0,0]^T$)  \\
   $\bm{b}$ & unite basis vectors of body-fixed frame($\bm{b}_1 =[1,0,0]^T$, $\bm{b}_2=[0,1,0]^T$, $\bm{b}_3=[0,0,1]^T$) \\ 
   $\zeta$  & rotation angle of the movable point about the x-axis of the body-fix frame\\
   $\bm{r}_s$ & position of the gravity centre of the rigid body in body-fixed frame\\
   $\bm{v}_s$ & linear velocity of the rigid body section in the body-fixed frame relative to the North-East frame observed in the body-fixed frame ($\bm{v}_s = \bm{v} + \bm{\omega}\times\bm{r}_s$).\\
   $\bm{\omega}_s$ & angular velocity of the rigid body section in the body-fixed frame relative to the North-East frame observed in the body-fixed frame($\bm{\omega}_s = \bm{\omega}$)\\
   $\bm{r}_p$ & position of the movable point in the body-fixed frame($\bm{r}_p = [r_p, -R_p\sin\zeta, R_p\cos\zeta ]$, where $R_p$ is deviation distance between the gravity centre of the movable point and x-axis of the body-fix frame.)\\
   $\bm{v}_p$ & linear velocity of the movable point with respect to the NED frame observed in the body-fixed frame($\bm{v}_p = \bm{v} +\bm{\omega}\times\bm{r}_p + \dot{\bm{r}}_p$) \\
   $\bm{\omega}_p$ & angular Velocity of the movable point with respect to the NED frame in the body-fixed frame($\bm{\omega}_p = \bm{\omega}+\dot{\zeta}\bm{b}_1$)\\
   $\bm{J}_s$ & inertial matrix of the rigid body \\
   $\bm{J}_p$ & inertial matrix of the movable point with respect to the body-fixed frame( $\bm{J}_p =^b\bm{R}_p \bm{J}^0_p$, where $^b\bm{R}_p$ and $\bm{J}^0_P$ is an initial inertial matrix, at this moment, $\zeta = 0$) 
    \begin{equation}
     ^b\bm{R}_p=
     \begin{bmatrix} 
      1 & 0 & 0 \\ 
      0 & cos\zeta & -sin\zeta \\
      0 & sin\zeta & cos\zeta  
     \end{bmatrix}
   \end{equation}
   \\ 
   \hline
   \end{tabularx}
\end{table}

\subsubsection{Extension of External Force and Moments Terms}

The last term in right side of (\ref{eq:kineticsEquations}) can be expanded as,

\begin{equation}
\begin{bmatrix}
\bm{F}_{ext}\\
\bm{T}_{ext}
\end{bmatrix}
=
\begin{bmatrix}
\Delta m(z)g {}^B\bm{R}_{I}\bm{i}_3 \\
(m_s\bm{r}_s + m_p\bm{r}_p + m_b\bm{r}_b)g \times {}^B\bm{R}_{I}\bm{i}_3
\end{bmatrix}
+
\begin{bmatrix}
\bm{F}_h\\
\bm{T}_h
\end{bmatrix}
+
\begin{bmatrix}
\bm{F}_m\\
\bm{T}_m
\end{bmatrix}.
\label{eq:externalTerm}
\end{equation}

In the above equation, $g=9.81$ is gravitational constant, ${}^I\bm{R}_B$ is a transformation matrix mapping vectors in the body-fixed frame into the inertial frame. The symbol $\bm{i}$ is the standard basis, which is $\bm{i}_1 = [1,0,0]^T$, $\bm{i}_2 = [0,1,0]^T$, $\bm{i}_3=[0,0,1]^T$. For the Earth's rotation is ignored, the NED frame and the inertial frame can be considered to be coincided. That means ${}^N\bm{R}_B = {}^I\bm{R}_B$, where ${}^N\bm{R}_B$ is a rotation matrix presenting the the pose of the body-fixed frame in the inertial frame, and also transforms the orientation of a vector with respect to the body-fixed frame into the NED frame.

The first term on right side of (\ref{eq:externalTerm}) is the buoyancy force and the buoyancy moment, which are,

\begin{align}
\Delta m(z)g = (m_b +m_s +m_p -\rho(z)(\frac{m_s + m_p}{p_5} - K_{vh}z))g,
\end{align}

\noindent where $z$ is the diving depth of the glider, $\rho(z) \approx density ~ of ~ deep seawater$, $p_5$ is the density of sea water. $K_{vh}$ is a coefficient of the volume variance according to different diving depths. For the sake of simplicity, the moment can be divided into three components: the rigid body section, the movable point, and the ballast, which are,

\begin{align*}
T_s = m_sg\bm{r}_s\times({}^B\bm{R}_N\bm{i}_3)\\
T_p = m_pg\bm{r}_p\times({}^B\bm{R}_N\bm{i}_3)\\
T_b = m_bg\bm{r}_b\times({}^B\bm{R}_N\bm{i}_3).
\end{align*}

The second term on the right side of (\ref{eq:externalTerm}) is referred to as the viscous hydrodynamic force and moment. It can be extended as, 

\begin{align}
\bm{F}_n &= {}^B\bm{R}_v
\begin{bmatrix}
-D \\
SF \\
-L 
\end{bmatrix} \\
\bm{T}_h &= {}^B\bm{R}_v
\begin{bmatrix}
T_{DL1}\\
T_{DL2}\\
T_{DL3}
\end{bmatrix}
\end{align}

\noindent where $D$, $SF$, and $L$ are the resistance, the side force and the lift force. The $T_{DL}$ is the hydrodynamic moment. The $\bm{R}^B_v$ is a transformation matrix mapping vectors from the velocity frame to the glider body-fixed frame. These are derived from hydrodynamic and hydrostatic coefficients, such as $K_{D0}$, $K_\beta$, $K_{L0}$, $K_L$, etc, through below equations,

\begin{equation}
    D=(K_{D0}+K_D\beta^2)V_r^2
\end{equation}

\begin{equation}
    SF=(K_\beta\beta)V_r^2
\end{equation}

\begin{equation}
    L=(K_{L0} + K_L\beta)V_r^2
\end{equation}

\begin{equation}
    T_{DL1}=(K_{MR}\beta + K_pp_r)V_r^2
\end{equation}

\begin{equation}
    T_{DL2}=(K_{M0} + K_M\beta K_qq_r)V_r^2
\end{equation}

\begin{equation}
    T_{DL3}=(K_{MY}\beta + K_rr_r)V_r^2
\end{equation}

\noindent where $p_r$, $q_r$, $r_r$ are elements in the angular velocity related to the fluid, hence the velocity vector is $\bm{\omega}_r = [p_r ~ q_r ~ r_r]^T$. The $V_r$ is the norm of linear velocity $\bm{v}_r = [\mu_r, \nu_r, \omega_r]$ related to the fluid, and the $\bm{R}^B_v$ is a transformation matrix mapping vectors from the velocity frame to the glider body-fixed frame, which are presented as,

\begin{align}
V_r = \sqrt{\mu^2_r+\nu^2_r+\omega^2_r}\\
\alpha = tan^{-1}\frac{\omega_r}{\mu_r}\\
\beta = sin^{-1}\frac{\nu_r}{V_r}
\end{align}

\begin{equation}
{}^B\bm{R}_v=
\begin{bmatrix}
\cos\alpha \cos\beta & -\cos\alpha \sin\beta & -\sin\alpha \\
\sin\beta &  \cos\beta & 0 \\
\sin\alpha \cos\beta & -\sin\alpha \sin\beta & \cos\alpha
\end{bmatrix}  
\end{equation}

The relative velocities $\bm{v}_r, \bm{\omega}_r$ are calculated from absolute velocities of the glider($\bm{v},\bm{\omega}$) and velocities of fluid currents ($\bm{v}_f,\bm{\omega}_f$).  

\begin{equation}
    \bm{v}_r = \bm{v} - {}^B\bm{R}_I \bm{v}_f
\end{equation}

\begin{equation}
    \bm{\omega}_r = \bm{\omega} - {}^B\bm{R}_I \bm{\omega}_f
\end{equation}

The third term on the right side of (\ref{eq:externalTerm}) is referred to as the inertial hydrodynamic force and moment, which is

\begin{align}
\begin{bmatrix}
\bm{F}_m\\
\bm{T}_m
\end{bmatrix}
=-\bm{M}_f
\begin{bmatrix}
\dot{\bm{v}}_r\\
\dot{\bm{\omega}}_r
\end{bmatrix}
\end{align}

\begin{align}
\bm{M}_f =
\begin{bmatrix}
\bm{M}_A & \bm{C}_A \\
\bm{C}_A^T & \bm{J}_A
\end{bmatrix} =
\begin{bmatrix}
\lambda_{11} & 0 & 0 & 0 & 0 & 0 \\
0 & \lambda_{22} & 0 & 0 & 0 & \lambda{26} \\
0 & 0 & \lambda_{33} & 0 & \lambda{35} & 0 \\
0 & 0 & 0 & \lambda_{44} & 0 & 0 \\
0 & 0 & \lambda_{53} & 0 & \lambda_{55} & 0 \\
0 & \lambda_{62} & 0 & 0 & 0 & \lambda_{66}
\end{bmatrix}
\end{align}

\noindent where, $\bm{M}_A$, $\bm{J}_A$ and $\bm{C}_A$ are the added mass matrix, the added inertial matrix, and cross terms $\lambda_{ii}$ are inertial hydrodynamic coefficients. 

\subsubsection{Variables Classification}

Since (\ref{eq:kineticsEquations}) involves dozens variables, these variables are categorized into different groups: constant variables, preset variables, states of the glider, control variables, iterative variables. Constant variables mean the variables in (\ref{eq:kineticsEquations}) are supposed to remain same all time. The constant values involve the standard basis and the gravity coefficient, i.e. $\bm{b}$, $\bm{b}_1 = [1,0,0]^T$, $\bm{b}_2=[0,1,0]^T$, $\bm{b}_3=[0,0,1]^T$, $\bm{i}$, $\bm{i}_1 = [1,0,0]^T$, $\bm{i}_2 = [0,1,0]^T$, $\bm{i}_3=[0,0,1]^T$, and $g=9.81$.

Preset variables are assigned during the simulation bootstrap and would remain same during the whole simulation. These variables and their descriptions are presented in Table \ref{tab:presetValues}. Generally, they are loaded through the URDF or Xacro robot description file.

 \begin{table}[H]
    \centering
    \caption{Preset Values}
    \label{tab:presetValues}
    \begin{tabularx}{0.8\textwidth}{c|X}
    \hline
    Name  & Description \\
     \hline
    $m_s$, $\bm{r}_s$  & mass and position vector of the gravity centre of rigid body section\\
    $m_p$ & mass of the movable point(battery package) \\
    $m_b$,$\bm{r}_b$ & mass and position vector of the gravity centre of the ballast \\
    $\bm{J}_0^p$ & initial inertial matrix for the movable point(battery package)\\
    $\bm{J}_S$ & initial matrix of the rigid-body section \\
    $R_p$ & deviation distance between the gravity centre of the movable point(battery package) and the axis-x of the body-fixed frame\\
    $p_5$, $K_{vh}$, $\rho$ & deep density, surface density, and deformation coefficients\\
    $K_{D0}, K_{D}, K_\beta, K_{L0}, K_L, K_{MR}, K_p, K_{M0}, K_M, K_q, K_{MY}, K_r, M_f$& hydrostatic coefficients \\
    $\Delta m_b$, $\Delta \zeta$, $\Delta r_{p1}$ & incremental values of actuators \\
    $Max~ r_{p1}, Min~ r_{p1}, Max~ Mb$ & ranges for control inputs\\
    \hline
    \end{tabularx}
\end{table}

States of a glider consist of the velocity, the orientation and position, as shown in Table \ref{tab:stateValues}. The states of the glider, solved from kinetic equations, are supposed to be with respect to the inertial frame which takes the Earth's rotation into account. Here, we ignore the Earth's rotation and assume that the states are with respect to the NED frame.

\begin{table}[H]
    \centering
    \caption{State Variables}
    \label{tab:stateValues}
    \begin{tabularx}{0.6\textwidth}{c|X}
    \hline
     Name  & Description \\
    \hline
    $\bm{v}$ & linear velocity of the body-fixed frame with respect to the NED frame observed in the body-fixed frame \\ 
    $\bm{\omega}$ & angular velocity of the body-fixed frame with respect to the NED frame observed in the body-fixed frame \\
    $\bm{q}$ & quaternion presenting the orientation of the body-fixed frame in the NED frame \\
    $\bm{p}$ & position vector of the body-fixed frame the in the NED frame \\
    \hline
    \end{tabularx}
\end{table}

Control variables indicate the variables which are directly impacted by low-level control inputs or commands. Commonly, they are the rotation angle of the battery $\zeta$, the translation of the battery $r_{p1}$, and the ballast mass $m_b$.

Iterative variables, depending on the states and the control variables, are ones which keep volatile among simulation iterations. In other words, due to variances of states and control variables, some associated variables in (\ref{eq:kineticsEquations}) would change correspondingly, which are referred to as iterative variables and shown in Table \ref{tab:updateingValues}. 

\begin{table}[H]
    \centering
    \caption{Iterative Values}
    \label{tab:updateingValues}
    \begin{tabularx}{0.7\textwidth}{c|X}
    \hline
    Name  & Dependencies \\
   \hline
   $\bm{v}_s, \bm{\omega}_s$ & depending on state variables $\bm{v}$ and $\bm{\omega}$, calculated by $\bf{\omega}_s= \bm{\omega}$, $\bm{v}_s=\bm{v}+\bm{\omega}\times\bm{r}_s$ \\
   $\bm{v}_p, \bm{\omega}_p$ & depending on the state variables ($\bm{v}$ and $\bm{\omega}$) and the gravity centre of the movable point($\bm{r}_p$), calculated by $\bm{v}_p = \bm{v} + \bm{\omega}\times\bm{r}_p + \dot{\bm{r}}_p$, $\bm{\omega}_p = \bm{\omega} + \dot{\bm{\zeta}}\bm{b}_1$ \\
   $\bm{v}_r$, $\bm{\omega}_r$ & depending on environmental current measurement and state values\\
   $\alpha$,$\beta$ & depending on the relative speed of fluid current $\bm{\omega}_r$ and $\bm{v}_r$ \\
   $\bm{R}^B_v$ & depending on $\alpha$ and $\beta$ \\
   ${}^B\bm{R}_N$ & depending on the orientation $\bm{q}$ \\
   $\bm{J}_p$ & depending on $\bm{R}^b_p$ \\
   $\bm{r}_p$ & depending on the translation of the battery package $r_{p1}$ and the rotation angle of the battery package $\zeta$ \\
   $\bm{R}^b_p$ & depending on the rotation angle of the battery package $\zeta$ \\
   \hline
    \end{tabularx}
\end{table}

In order to reduce redundant computations in simulations, approximations for some variables are conducted, as shown in Table \ref{tab:approximation}. These approximations are mainly for some trivial derivatives and second-order derivatives.

\begin{table}[H]
    \centering
    \caption{Approximation}
    \label{tab:approximation}
    \begin{tabularx}{0.4\textwidth}{c|X}
    \hline
    Name  & Description \\
    \hline
    $\dot{\bf{\omega}}_r$ & $\dot{\bf{\omega}}_r= 0$ \\
    $\dot{\bf{v}}_r$ & $\dot{\bf{v}}_r= 0 $ \\
    $\dot{\zeta}$ & $\dot{\zeta}=\Delta\zeta$ or $0$ if the battery package rotated to the desired angle\\
    $\ddot{\zeta}$ & $\ddot{\zeta}=0$ \\
    $\dot{\bf{r}}_p$ & $\dot{\bf{r}}_p =[\Delta r_p1, -R_pcos\zeta \dot{\zeta}, -R_psin\zeta \dot{\zeta}]^T$ \\
    $\ddot{\bf{r}}_p$ & 0\\
    \hline
    \end{tabularx}
\end{table}

\subsubsection{Control Variables Variation}

Since restrictions of electronic or mechanical structures, the control variables can't reach the given control inputs immediately. In realistic situations, the control variables descend or ascend gradually to approach the control inputs given from an operator or a high-level control algorithm, through relevant electronic or mechanical structures regulating the pose of the battery and the ballast mass. In this case, the control input can be regarded as the desired control values $Du$($D\zeta, Dr_{p1}, Dm_b$), and incremental value(incremental value per unit time) for the control variables are $\Delta u$, where $u$ can be substituted with $\zeta$, $r_{p1}$, and $m_b$. To mimic the work pattern of electronic or mechanical structures, variation of control variables will obey Algorithm \ref{al:controlVariable}.

\begin{algorithm}[H]
\caption{Control Variable Variation}
\label{al:controlVariable}
\begin{algorithmic}
    \State $Du$ Desired Control value
\State $u$ Control variable
\State $\Delta u$ Incremental value per unit time for the control variables
\State $\Delta t$ Endurance of one simulation iteration 
\If{$u_d - u > \Delta u  \Delta t $}
    \State $u= u +  \Delta u \Delta t$
\ElsIf{$u_d - u < \Delta u \Delta t$}
    \State $u= u -  \Delta u \Delta t$
\Else 
    \State $u=u_d$
\EndIf
\end{algorithmic}
\end{algorithm}

\subsection{Coordinates Transformation}

\begin{figure}[H]
    \centering
    \includegraphics[width=0.45\linewidth]{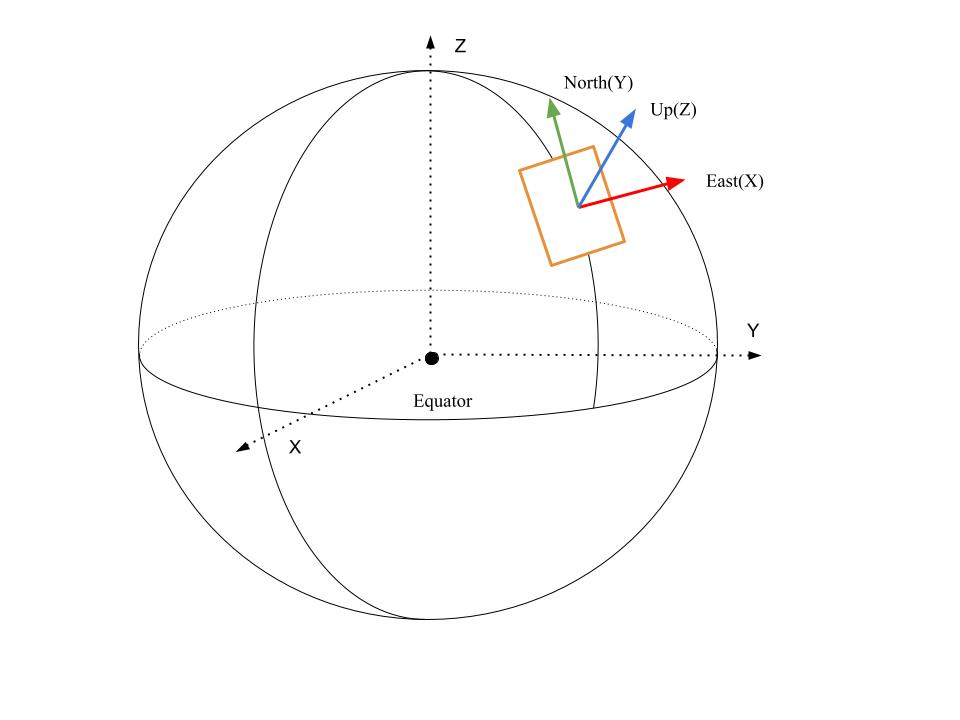}
    \includegraphics[width=0.45\linewidth]{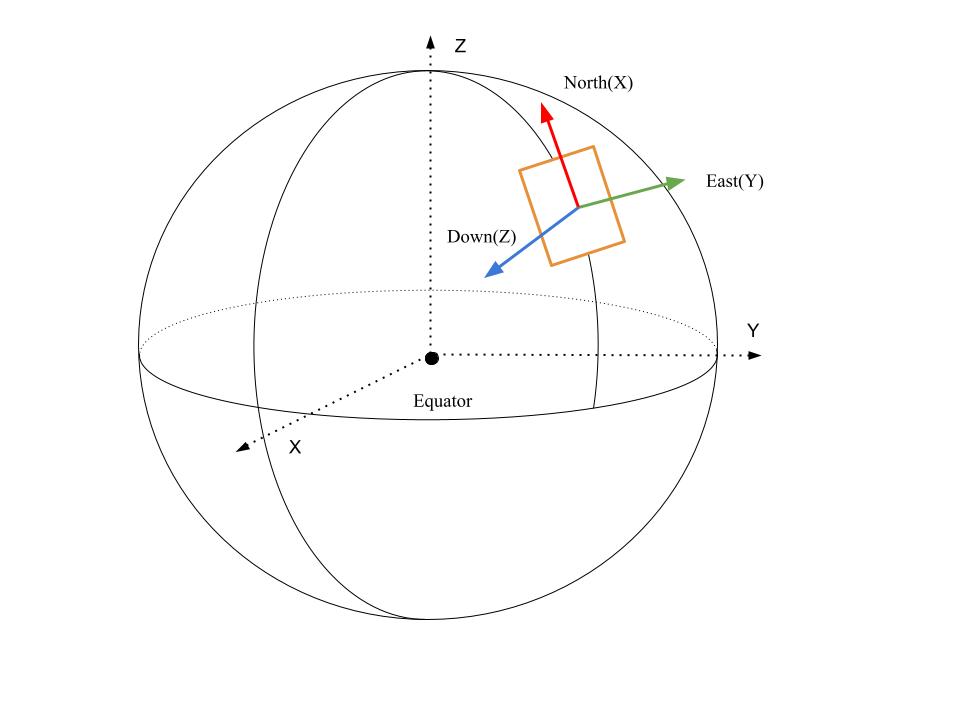}
    \caption{Comparison of ENU and NED Frames: Left figure presents an ENU Frame; Right figure presents a NED frame.}
    \label{fig:ENUandNED}
\end{figure}

The kinetic equations (\ref{eq:kineticsEquations}) are derived in the NED frame. Hence, the states(pose, velocities, and accelerations) of a glider solved from these differential equations are presented in the NED frame as well. However, the states of the glider model in the Gazebo simulation are described in the ENU frame \cite{gazeboCoordinates}. That means the state variables are supposed to be transformed into the ENU frame before updating the accelerations of a glider in the Gazebo world via model plugin APIs. These two frames are referred to as world frames or global frames, which are depicted in Fig.\ref{fig:ENUandNED}. 

\begin{figure}[H]
    \centering
    \includegraphics[width=0.7\linewidth]{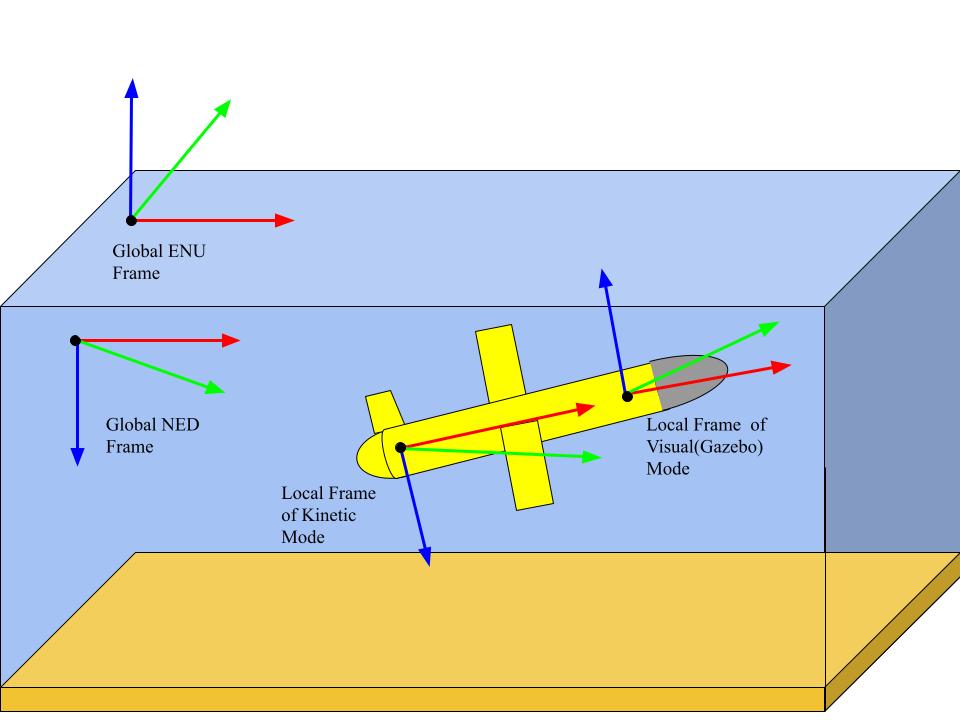}
    \caption{Frames in the simulator. Note that the two global frames and two local frames share the same original points, which means the distances of between the two local frames and the two global frames don't exist actually. We separate these frames so that they can be recognized visually. }
    \label{fig:frameDescription}
\end{figure}

Furthermore, local frames or body-fixed frames of the Gazebo model(visual model) and the kinetic model are different as well. Because the local frame usually is parallel to its world frame. The local frame of the Gazebo model(visual model) comply the ENU convention, while that of kinetic model obeys the NED convention. Similar with world frames, they share the same original point. The relations among global frames and local frames are illustrated in Fig.\ref{fig:frameDescription}.

Transformations from the NED frame to the ENU frame are categorised into following conditions. In this part, the symbol $k$ indicates the kinetic model's local frame, 
and $v$ represents the Gazebo(visual) model's local frame. The NED frame is abbreviated by $N$, and $E$ means the ENU frame.  
 
\begin{itemize}
    \item Orientation Transformation(${}^N\bm{R}_{k} \rightarrow ~ {}^E\bm{R}_{v}$) 
    
    Given the orientation of the kinetic model in the NED frame ${}^N\bm{R}_{k}$, obtain the orientation of Gazebo model(visual model) in the ENU frame ${}^E\bm{R}_{v}$:

    \begin{equation}
        {}^E\bm{R}_{v}  = {}^E\bm{R}_{N} ~ ^N\bm{R}_{k} ~ ^{k}\bm{R}_{v}
    \end{equation}

    \item Position Transformation($\bm{p}^N_{k} \rightarrow \bm{p}^E_{k}$)
    
    Given a vector of position of the kinetic model in NED frame $\bm{p}^N_{k}$, obtain the position vector of the Gazebo model(visual model) in the ENU frame $\bm{p}^E_{k}$:

    \begin{equation}
        \bm{p}^E_{k} = ~^E\bm{R}_ {N} ~ \bm{p}^N_{k}
    \end{equation}
   
   We can conclude $|\bm{p}|^N_{k} = |\bm{p}|^E_{v}$, since two local coordinates' origins are coincided. 

   \item  Linear Velocity, Acceleration and Rotation Rates Transformation($\bm{v}^k_ {Nk} \rightarrow \bm{v}^E_{Ev}$) 
   
   Given linear or rotation velocities of kinetic model $\bm{v}^{k}_ {Nk}$ in NED frame,
   obtain linear or rotation velocities $\bm{v}^{E}_{E v}$ of the Gazebo model(visual model) in the ENU frame: 
   
   \begin{equation}
       \bm{v}^{E}_ {E v} = ^{E} \bm{R}_ {k} (\bm{v}^{k}_ {EN} + \bm{v}^{k}_ {Nk} + \bm{v}^{k}_ {k v})
   \end{equation}

   Since the world frame and the NED frame are static, and local frames $k$ and $v$ have zeros relative linear velocities, $\bm{v}^{k}_ {k v} = 0$ and $\bm{v}^{k}_ {EN} = 0$. Therefore, $\bm{v}^{E}_ {E v} =^E\bm{R}_ N ~ ^N\bm{R}_ {k} (\bm{v}_ {N k}^{k})$. Similar deductions can be made for accelerations and rotation rates $\bm{\omega}$(the $\bm{v}^{k}_ {k v}$ can further be decomposed into addition of velocities in inertial frame).
\end{itemize}

In the implementation, $^w\bm{R}_N$ and $^{k}\bm{R}_{v}$ are constants and can be calculated through the Euler angles,

\begin{equation}
      ^w \bm{R}_ N = \bm{R}_Z(\frac{\pi}{2})\bm{R}_Y(\pi)\bm{R} _X(0)
\end{equation}

\begin{equation}
    ^{k} \bm{R}_ {v} = \bm{R} _Z(0) \bm{R} _Y(0) \bm{R} _X(\pi).
\end{equation}

The rotation matrix $^N\bm{R}_{k}$ is obtained by solving kinematics equations. The reverse transformations can be realised through the matrix reverse operation.


\section{LQR Control Module}

The control of buoyancy-driven glider is implemented by the LQR method. The controller receiving the desired gliding velocity, pitch and heading angles from the guidance system or an operator, and then generates control commands for low-level actuators to tune the craft to reach these desired states. 

There are various control strategies for buoyancy-driven gliders have been reported in recent years\cite{ullah2015underwater}\cite{wu2022feedback}. Among the various controllers, PID (proportional–integral–derivative controller) and LQR(Linear Quadratic Regulator) are two most commonly implemented methods. Compared to the PID controller, the LQR is a standard linear optimal control design method which produces a stabilizing control law that minimizes a cost function that is a weighted sum of the squares of the states and input variables\cite{leonard2001model}. It is capable of finding optimal control inputs that minimise overall system performance criteria\cite{song2023evaluation}. In many scenarios, researchers have presented that the LQR outperforms the PID controller in glider systems\cite{noh2011depth}\cite{leonard2001model}.

\subsection{LQR Control Strategy}


When the glider floats up, the guidance system received the signal from GNSS(Global Navigation Satellite System), correct its location. Then it would send instructions to the LQR Controller. Once LQR gains are solved, the controller would send low-level commands to the actuators. Note that LQR controllers are decoupled into the vertical plane and the horizontal plane. The state-space representations linearized from (\ref{eq:kineticsEquations}) at equilibrium points are different in ascending and descending processes. Hence, there are totally four different LQR gains are supposed to be obtained, prior to start a work circle. They individually match four different circumstances: Ascending in the horizontal plane; Descending in the horizontal plane; Ascending in the vertical plane; Descending in the horizontal plane.

\begin{figure}[H]
    \centering
    \includegraphics[width=1\linewidth]{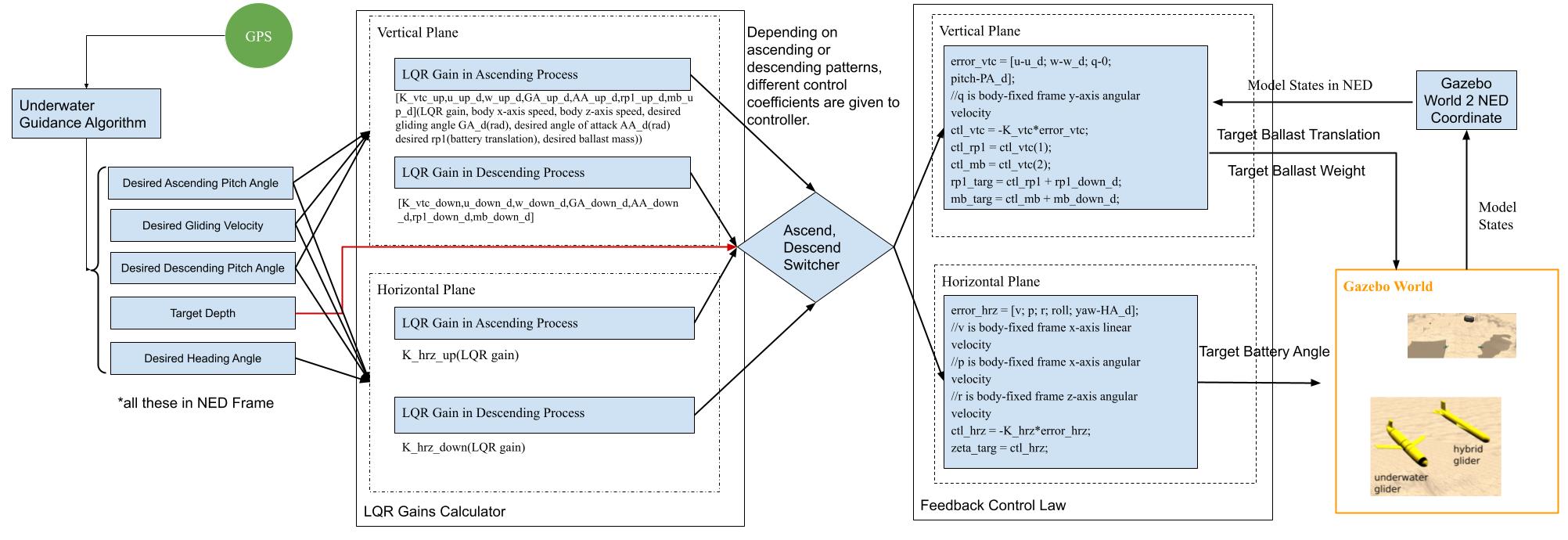}
    \caption{LQR Control Strategy}
    \label{fig:LQRcontroller}
\end{figure}

The control strategy is shown in Fig.\ref{fig:LQRcontroller}. The guidance system provides the desired pitch angles in ascending and descending processes, the desired heading angle, and the desired velocity for the LQR controller. Before the glider executes next work circle shown in Fig.\ref{fig:pitchAngle}, the four LQR control gains are computed. Once computations complete, the feedback control laws in horizontal and vertical planes are implemented. States of the glider, including orientations and velocities are transferred into the NED frame, are fed into the control laws. The outputs of the controller are commands of the battery linear translate, the ballast mass, and the battery angle, provided for the low-level actuators. The glider's orientations and velocities can be obtained through a DVL sensor or the Gazebo model states publisher. However, sometimes, in order to minimize power consumptions, a DVL might be not equipped. In this case, we can set the velocity error zero in the feedback control laws for detaching the speed control.


\subsection{LQR Controller Derivation}

The LQR controller has been reported in \cite{leonard2001model}\cite{song2023evaluation}\cite{zhang2013spiraling}. Here, a short summary of LQR controller derivation is presented.
The LQR control problem can be defined to minimize the cost function,

\begin{equation}
    J = \int^\infty _0 \Delta \bm{x}^T\bm{Q}\Delta \bm{x} + \Delta \bm{u}^T \bm{R} \Delta \bm{u} ~ dt
    \label{eq:LQRcost}
\end{equation}

\noindent where $\bm{Q}$ and $\bm{R}$ are state and control penalty matrices. The matrices $\bm{Q}$ and $\bm{R}$ were chosen to ensure well-behaved dynamics and to prevent large motions in the movable mass position and variable mass that would exceed physical limitations. The weight selections are given by \cite{leonard2001model} \cite{song2023evaluation}. Let $\Delta x$ and $\Delta u$ be the small disturbances of state variables and control inputs related to the glider equilibrium, which are represented as,

\begin{align*}
    \Delta \bm{x} = \bm{x} - \bm{x}_{eq} \\
    \Delta \bm{u} = \bm{u} - \bm{u}_{eq}
\end{align*}
\noindent where $\bm{x}$ and $\bm{u}$ are states variables and control inputs respectively. We denote with subscript $eq$ variables at the glider equilibrium point.

In order to solve (\ref{eq:LQRcost}), the kinetic equations (\ref{eq:kineticsEquations}) should be linearized to the form of, 

\begin{equation}
    \Delta \dot{\bm{x}} = \bm{A} \Delta x + \bm{B} \Delta \bm{u}, 
\end{equation}

\noindent where $\bm{A}$ is the state matrix and $\bm{B}$ is the input matrix. 

The corresponding control law is $\bm{u} = - \bm{K}\Delta \bm{x}$ where $\bm{K}$ is computed using the Python Control Systems Library\cite{pcsl} and the SciPy\cite{scipy} from the Riccati equations formed by $\bm{A}$, $\bm{B}$, $\bm{Q}$, and $\bm{R}$. In this case, control inputs are $\bm{u}= [r_{p1}, m_b, \xi] $, $r_{p1}$ the battery package translation, $m_b$ the ballast mass, and $\xi$ the rotation angle of battery package. The state variables are $x=[u,w,q,\theta]$, where $u$ is the forward velocity, $w$ is the heave velocity, $q$ is the pitch angle, and $\theta$ is the yaw angle or the heading angle. 

\section{Recursive Guidance Moule}

The aim of the recursive guidance strategy is to suit the long-term operation, the low manoeuvrability, and the unsatisfactory accuracy of underwater localisation technologies. The former two features come from the special structure of this class of vehicles, as mentioned in Section \ref{sec:intro}. The latter feature results from the fact that this kind of gliders usually do not equip with the DVLs(Doppler Velocity Logs) or other acoustic positioning devices in order to minimise the power cost. Because of their sawtooth travel pattern, the GNSS is a common means of relocation as soon as they float up the sea surface. 

Based on common realistic scenarios, a guidance strategy is developed based on the waypoint system. The selected waypoints(latitudes and longitudes) and target depths are preloaded via a task file. They are stored in a database and used for generation of a trajectory or a path for the moving craft to follow \cite{fossen2011handbook}.  With these information, the guidance would calculate the desired heading, the pitch angles when the vehicle is on the sea surface. Every floating up is regarded as an iteration for the proposed guidance algorithm, with updating latest position information via GNSS(or GPS), the desired heading and pitch angles are corrected. In this case, pitch angles are changing, and become steeper and steeper as the glider approaching target positions. Alternatively, the pitch angles during a whole mission can be assigned from the task file. In this case, the pitch angles are independent of the guidance system.

\begin{figure}[H]
    \centering
    \includegraphics[width=0.8\linewidth]{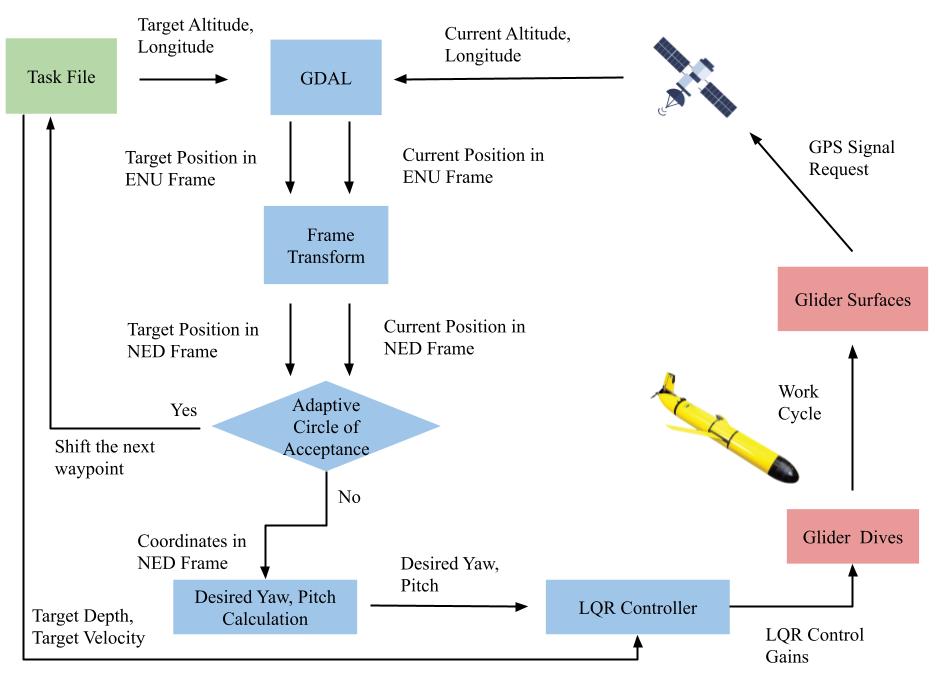}
    \caption{Recursive Guidance System}
    \label{fig:guidanceStrategy}
\end{figure}

Fig.\ref{fig:guidanceStrategy} presents the whole guidance strategy which can be divided into two subprocesses. The GPS sensor produces the latitude and longitude coordinates, which are supposed to be converted into the NED frame. According to given latitude and longitude from a task file, the desired heading and pitch angles are computed, or pitch angles are pre-assigned.   

\subsection{Conversion from Geographic Coordinate System to NED Frame}

The simulator fully inherits the sensor plugins from the DAVE project. The GPS module developed in the DAVE project transforms ENU coordinates to the WGS84 geodetic system(latitude \& longitude system)\cite{zhang2022dave}\cite{gazeboCoordinates}, rather than directly acquiring the latitude \& longitude through the Gazebo server. It is implemented by specifying certain EPSG(European Petroleum Survey Group) codes through the GDAL(Geospatial Data Abstraction Library) Warp API, which provides services for the high performance image warping using the application provided geometric transformation functions\cite{gdal}. In the simulator, we inverse the transformation by using the GDAL Warp API to obtain the coordinates in the ENU frame. Then, these position vectors are transformed into the NED frame. 

\subsection{Desired Heading Angle Identification}

The desired heading angle(Euler angles) is determined by Line-Of-Sight(LOS) steering law\cite{fossen2011handbook}. The LOS vector is defined by the glider's position and the target(the waypoint). Both of them are in the NED frame.

Firstly, the LOS vector between the current position $\bm{p}_t$ and the target position $\bm{p}_k$ is defined as,

\begin{equation}
    \bm{p}_d = \bm{p}_t - \bm{p}_{k}
    \label{eq:losP}
\end{equation}

\noindent where $\bm{p}_d$ is a LOS vector between the current position to target position. The $\bm{e}_d$ is the normalised vector of $\bm{p}_d$,

\begin{equation}
    \bm{e}_d = \frac{\bm{p}_d}{norm(\bm{p}_d)}.
\end{equation}

The cosine of the desired heading angle is the inner product of $\bm{e}_d$ and the x-axis of the NED coordinate,

\begin{equation}
    \cos(\theta) = \bm{e}_d^T \bm{e}_x
\end{equation}

Since $\bm{e}_x = [1,0,0]^T$, 

\begin{equation}
    \cos(\theta) = \bm{e}_d^1
\end{equation}
\noindent where $\bm{e}_d^1$ is the first component of the vector $\bm{e}_d$.

The heading angle $\theta$ is in the range of $[-\pi, \pi]$, and is computed by $\theta =\cos^{-1}(\bm{e}_d^1)$. However, the return value of $\cos^{-1}()$ function lies between $0$ and $\pi$.
The sign of $\theta$ can be determined by the second component of the distant vector, which is $\bm{e}_d^2$. 
When $\bm{e}_d^2 > 0$, the sign is $-1$. Oppositely, the sign is $1$, when $\bm{e} _d^2<0$.
(If $\bm{e}_d^2 = 0$, the $\bm{e}_d^1 =1$, and $\theta$ is zero.) 

\subsection{Desired Pitch Angle Identification}

\begin{figure}[H]
    \centering
    \includegraphics[width=0.6\linewidth]{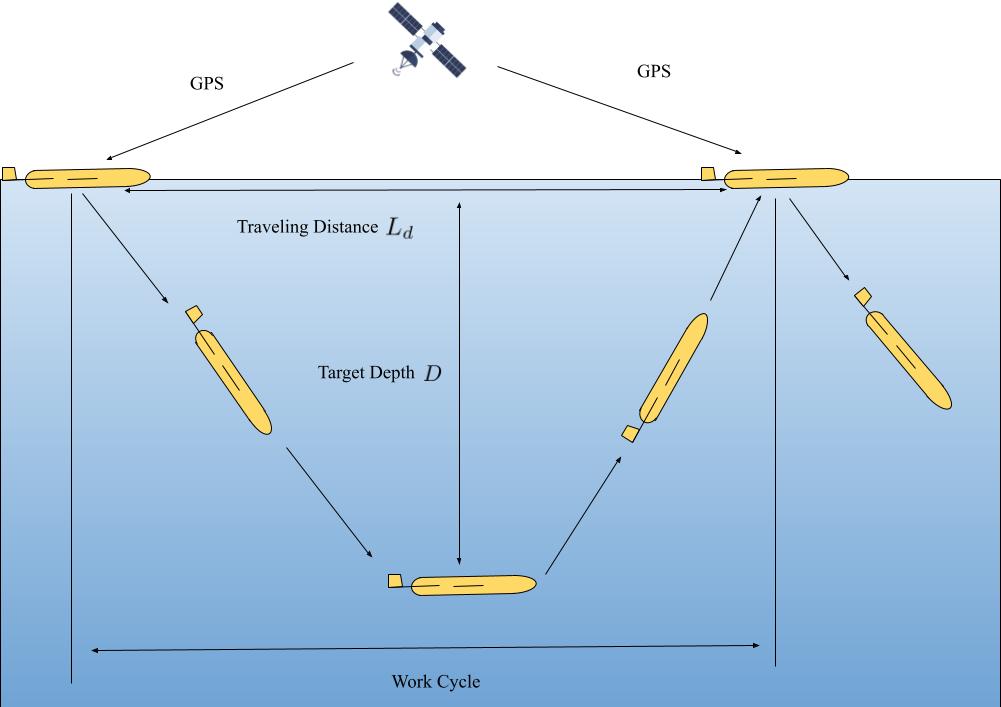}
    \caption{Pitch Angle Identity}
    \label{fig:pitchAngle}
\end{figure}

Once pitch angles are determined by the guidance system, the magnitude of pitch angles($\phi$) in ascending and descending processes are assumed to be identical. It is determined by the expected travelling distance($L$) of one work cycle and the target depth($D$), as shown in Fig.\ref{fig:pitchAngle},

\begin{equation}
    \phi = \tan^{-1}(\frac{L_d}{2D})
\end{equation}

The travelling distance($L_d$) is the quotient of the distance($L_t$) from the current position to the target position divided by the minimum loops number, i.e. $L_d = \frac{L_t}{minimum ~ loops ~ numbder}$. As the glider tends to approach the target, the pitch angle will become steeper and steeper, and finally hits the boundary of the range of pitch angle. By this way, a glider is expected to get the target as closely as possible. 

\subsection{Adaptive Circle of Acceptance}  

When moving along a piece wise linear path made up of n straight-line segments connected by $n+1$ waypoints, a switching mechanism for selecting the next waypoint is needed. This waypoint($\bm{p}_{k+1} = (x_{k+1}$, $y_{k+1})$) can be selected on the basis of whether or not the craft lies within a circle of acceptance with the radius $R$ around the current waypoint($\bm{p}_k = (x_{k+1}$, $y_{k+1})$). Moreover, if craft positions $\bm{p}_t$ at time t satisfy,

\begin{equation}
    |\bm{p}_t - \bm{p}_k|^2 \leq R^2
    \label{eq:coa}
\end{equation}
or,
\begin{equation}
    |\bm{p}_d| \leq R
\end{equation}

\noindent the next waypoint ($\bm{p}_{k+1}$) should be selected. 

For normal surface ships and submarines, the circle of acceptance $R$ is constant. However, the buoyancy-driven gliders cannot be located precisely  when it is submerged. Hence, it might pass by the waypoint or the target when it is dividing in underwater. An adaptive circle of acceptance for buoyancy-driven gliders is defined as,

\begin{equation}
    R_a = \sqrt{|\bm{p}_{d}|^2-|\bm{p}_d \cos{\theta}|^2}
\end{equation}

\noindent where $\bm{p}_d$ is the LOS vector($\bm{p}_d = |\bm{p}_t - \bm{p}_k| $) and,

\begin{equation}
    \cos{\theta} = \frac{\bm{p}_d(\bm{p}_t - \bm{p}_{t-1})}{|\bm{p}_d||\bm{p}_t - \bm{p}_{t-1}|}
\end{equation}

\noindent If $\cos{\theta} \geq 0$, the current position $\bm{p}_t$ and the previous position $\bm{p}_{t-1}$ are on the either side of the waypoint. It implies the vehicle past the waypoint $\bm{p}_k$ and the adaptive circle of acceptance should be derived and checked. Oppositely, we will not process further steps and let the vehicle continue moving. when $\cos{\theta} = 1$, the adaptive circle of acceptance can be simplified to,

\begin{equation}
    R_a = L - |\bm{p}_d|
\end{equation}

\noindent where $L$ is the horizontal distance of the previous work cycle, which $L=\frac{2D}{\tan\phi_d}$, $\phi_d$ is the desired pitch angle for the previous work cycle, $D$ is the depth of the previous work cycle. In realistic situations, the simplified adaptive circle of acceptance is preferred, for $\theta \rightarrow 0$ as the glider approaches the waypoint.

When the glider finishes the next work cycle and floats up, the guidance system detects if the adaptive circle of acceptance derived by the current position $\bm{p}_{t}$ satisfies,

\begin{equation}
        R_a^2 \leq R^2
\end{equation}

\noindent that means the glider has past by the target within the constant acceptable circle $R$, and the next waypoint should be shifted.

\section{Manoeuvrability Check}

Due to the low manoeuvrability of buoyancy-driven gliders, a tool is designed to predict if the buoyancy-driven glider is capable to reach the target or the next waypoint in a single turning. The tool is developed for helping to generate waypoints for the guidance system. The initial heading angle, the forward velocity, the initial position, and the target position are identified before the prediction. Then the trajectory of the glider with least turning radius is estimated. Subsequently, the intersection point of the least turning radius trajectory and a line defined by the initial position and target position(waypoint) in x-y plane, is located.

\subsubsection{Trajectory Prediction of Least Turning Radius}

When the glider is making a turning, its positions can be estimated by, 

\begin{equation}
    \begin{bmatrix}
x \\
y
\end{bmatrix} =\begin{bmatrix}
\int_0^T \cos(\theta_i + rt)v dt \\
\int_0^T \sin(\theta_i + rt)v dt
\end{bmatrix}
\end{equation}

\noindent where $r$ is the rotation rate, $v$ is the forward velocity, and $\theta_i$ is the initial heading angle.  

After integrating, the positions at the time $T$ is,

\begin{equation}
    \begin{bmatrix}
x \\
y
\end{bmatrix} =
\begin{bmatrix}
\frac{1}{r}\sin(\theta_i + rT)v  - \frac{1}{r} \sin(\theta_i)v \\
-\frac{1}{r}\cos(\theta_i + rT)v + \frac{1}{r} \cos(\theta_i)v
\end{bmatrix}.
\label{eq:positionEs}
\end{equation}

If the lower bound of the forward velocity substitutes the forward velocity and the upper bound of rotation rate substitutes the rotation rate in (\ref{eq:positionEs}), the estimated positions will satisfy the least turning radius, which is proved by,
\begin{equation}
    r_{upper}, v_{lower} = arg min_{r,v}||f(r, v)|| = arg min_{r,v}|\begin{bmatrix}
\frac{1}{r}\sin(\theta_i + rT)v  - \frac{1}{r} \sin(\theta_i)v \\
-\frac{1}{r}\cos(\theta_i + rT)v + \frac{1}{r} \cos(\theta_i)v
\end{bmatrix}|
\end{equation}

\noindent for an arbitrary time $T$. 


The sign of rotation rate is able to let the glider reach the desired heading angle along the shortest path. To implement it, we need two intermediate variables: the angle deviation($\theta_e$) and the optimal angle deviation($\theta_o$). The $\theta_e$ is defined as,

\begin{equation}
    \theta_e = \theta_d - \theta_c,
\end{equation}

\noindent where $\theta_c$ is the current heading angle, and $\theta_d$ is the direction angle of a vector between the current waypoint($\bm{p}_k$) and the next waypoint($\bm{p}_{k+1}$), which can be solved by,

\begin{equation}
    \bm{p}_{k,k+1} = \bm{p}_{k+1} - \bm{p}_k;
    \label{eq:directV}
\end{equation}

\begin{equation}
    \theta_d = atan2(\bm{p}_{k,k+1})
\end{equation}

\noindent where $atan2(y,x)$ is the four-quadrant version of $arctan(y/x) \in [-\pi/2, \pi/2]$.

The value of the optimal angle deviation ($\theta_o$) is dependent on $\theta_e$,

\begin{itemize}
    \item If $\theta_e < \pi$, then $\theta_o = \theta_e$. 

    \item If $\theta_e > \pi$, then $\theta_o = \theta_c + \theta_d - 2\pi$. 
\end{itemize}

\noindent When $\theta_o > 0$, the sign of the rotation rate $r$ is positive, otherwise the sign of the rotation rate $r$ is negative.

\subsubsection{Intersection Location}

\begin{figure}[H]
    \centering
    \includegraphics[width=0.7\linewidth]{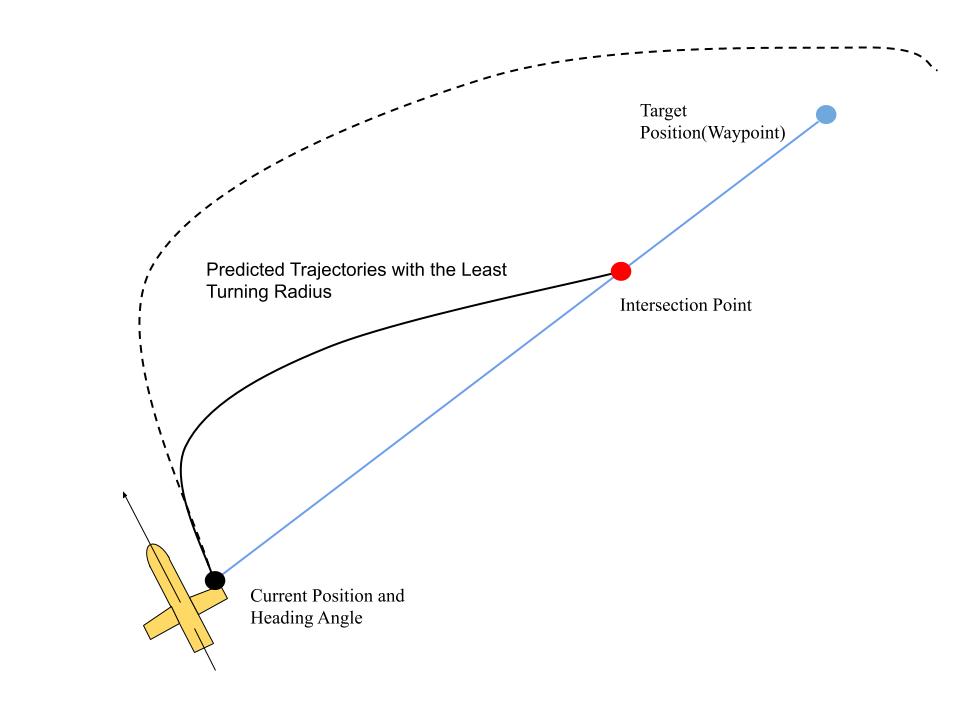}
    \caption{Intersection Location: The black dotted line indicates the glider cannot reach the target within a single turning theoretically; The black solid line shows the glider could arrive at the target without external disturbances}
    \label{fig:intersectionLocation}
\end{figure}

A intersection point between the vector $\bm{p}_k$ defined in (\ref{eq:directV}) and the estimated trajectory of least turning radius (\ref{eq:positionEs}) can be found by solving linear equations. If the intersected point locates within the interval between initial position and target position, it means without any disturbance, the glider could reach the target point from the initial position, as the black solid line shown in Fig.\ref{fig:intersectionLocation}; Otherwise, it is impossible for the glider to arrive at the target position, as the black dotted line presented in Fig.\ref{fig:intersectionLocation}.

\section{Example: Petrel-II glider}

Petrel II glider is a 1.8m-long torpedo-shaped buoyancy driven glider, developed by Tianjin University independently. It has passed a 1500m-deep water test in the northern part of the South China Sea\cite{liu2018using}. The configuration information and hydro-parameters of the gilder have been published\cite{wu2021optimization}.

\begin{figure}[H]
    \centering
    \includegraphics[width=0.8\linewidth]{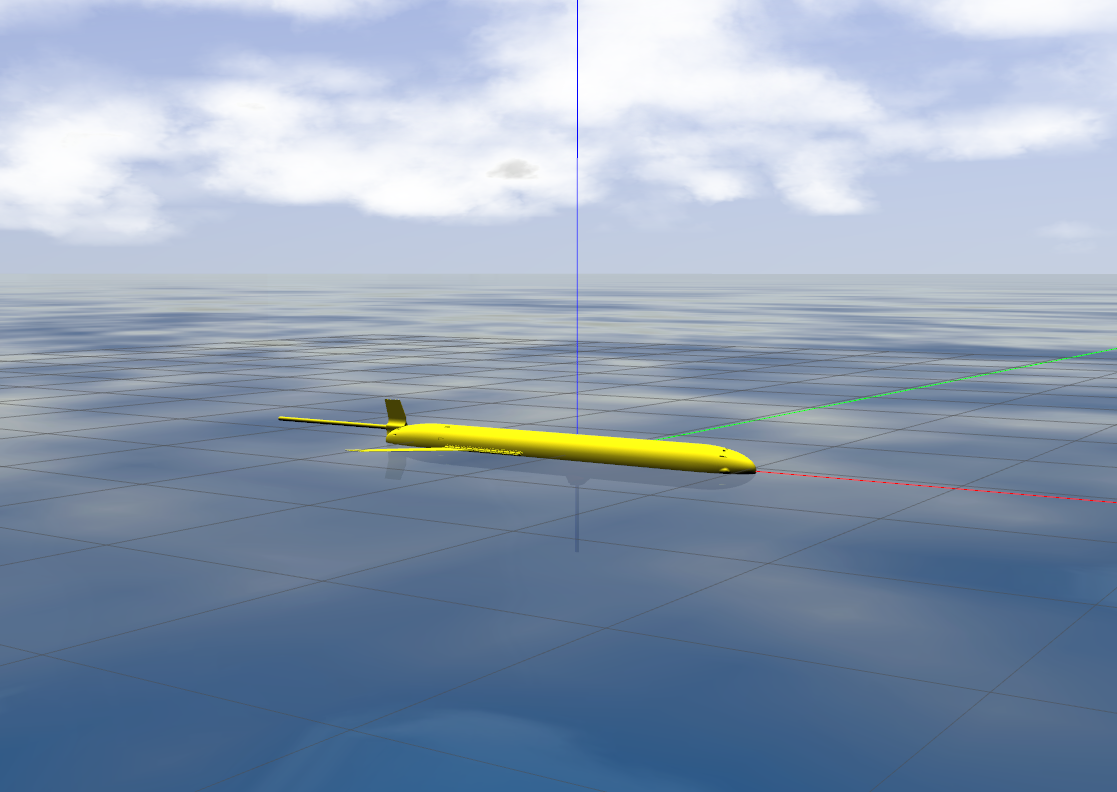}
    \caption{Floating of the glider in the Gazebo World}
    \label{fig:floating}
\end{figure}
\begin{figure}[H]
    \centering
    \includegraphics[width=0.8\linewidth]{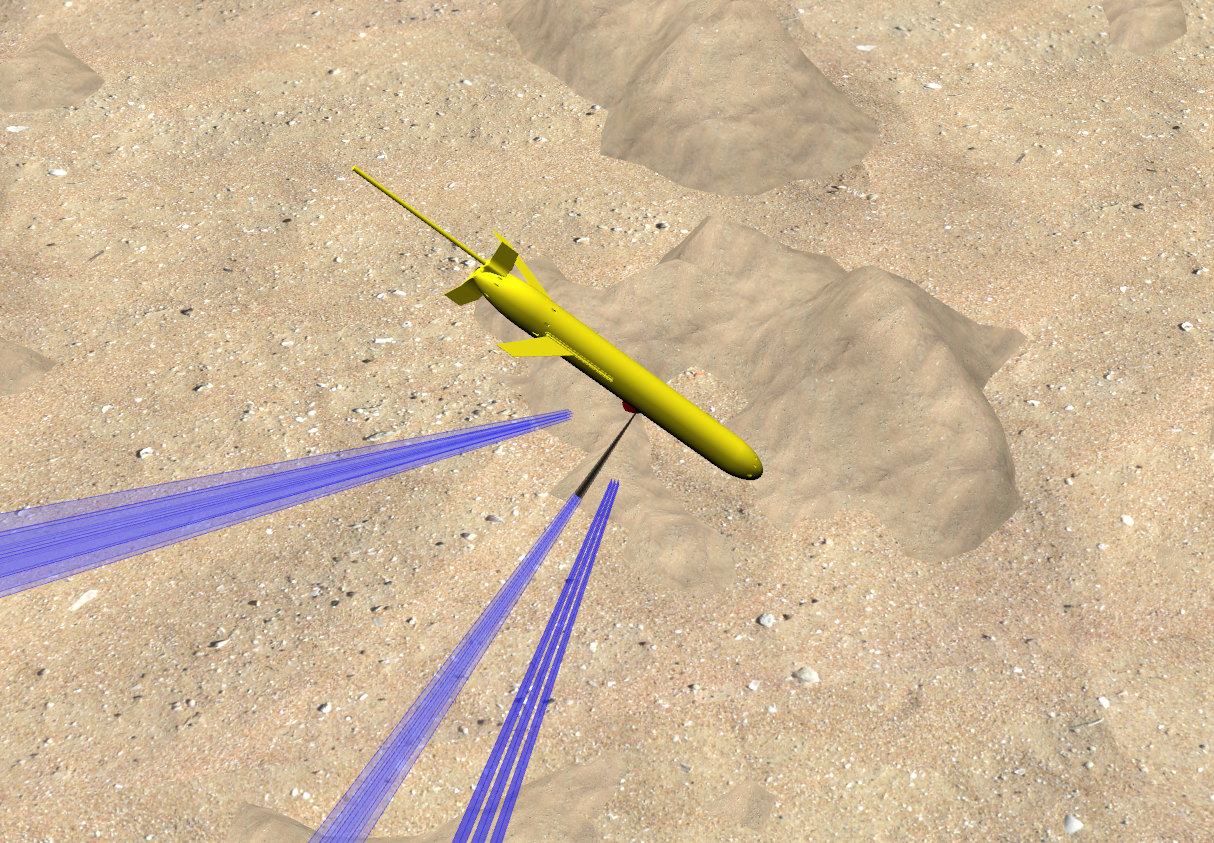}
    \caption{Descending of the glider in the Gazebo World}
    \label{fig:descendingGlider}
\end{figure}
\begin{figure}[H]
    \centering
    \includegraphics[width=0.8\linewidth]{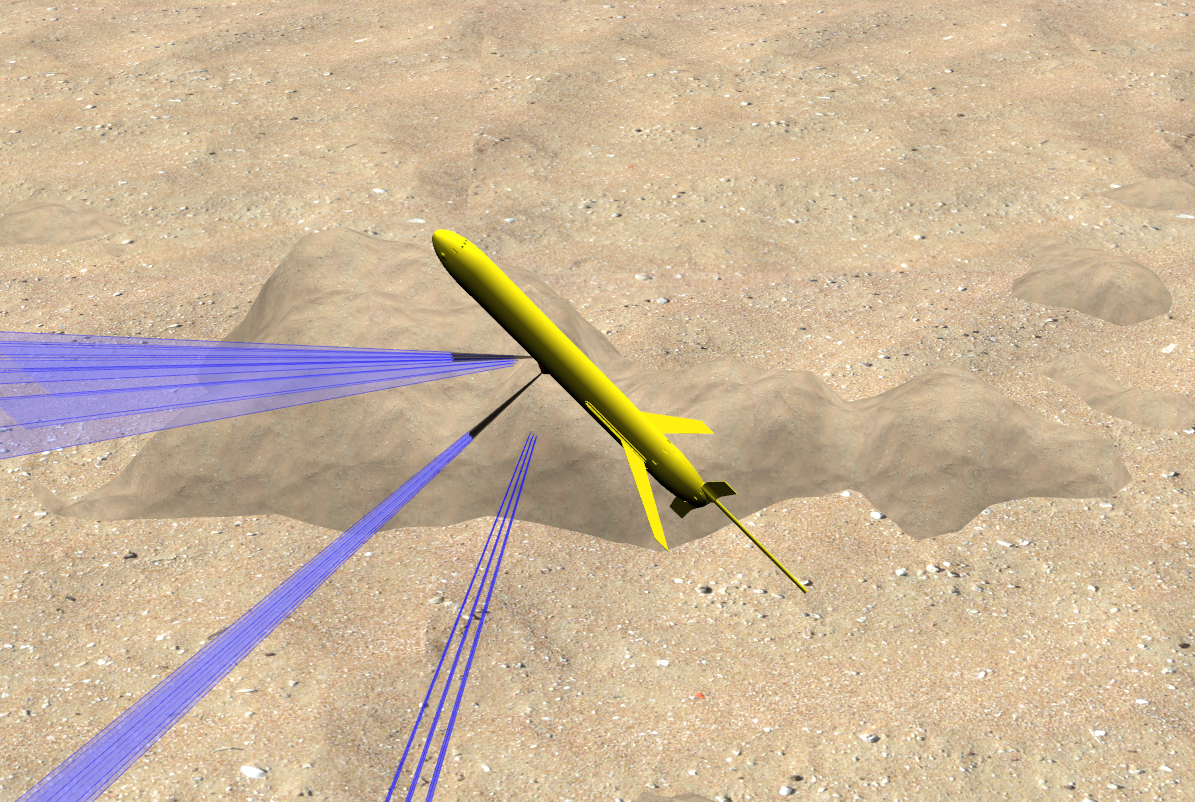}
    \caption{Ascending of the glider in the Gazebo World}
    \label{fig:ascendingGlider}
\end{figure}
\begin{figure}[H]
    \centering
    \includegraphics[width=0.8\linewidth]{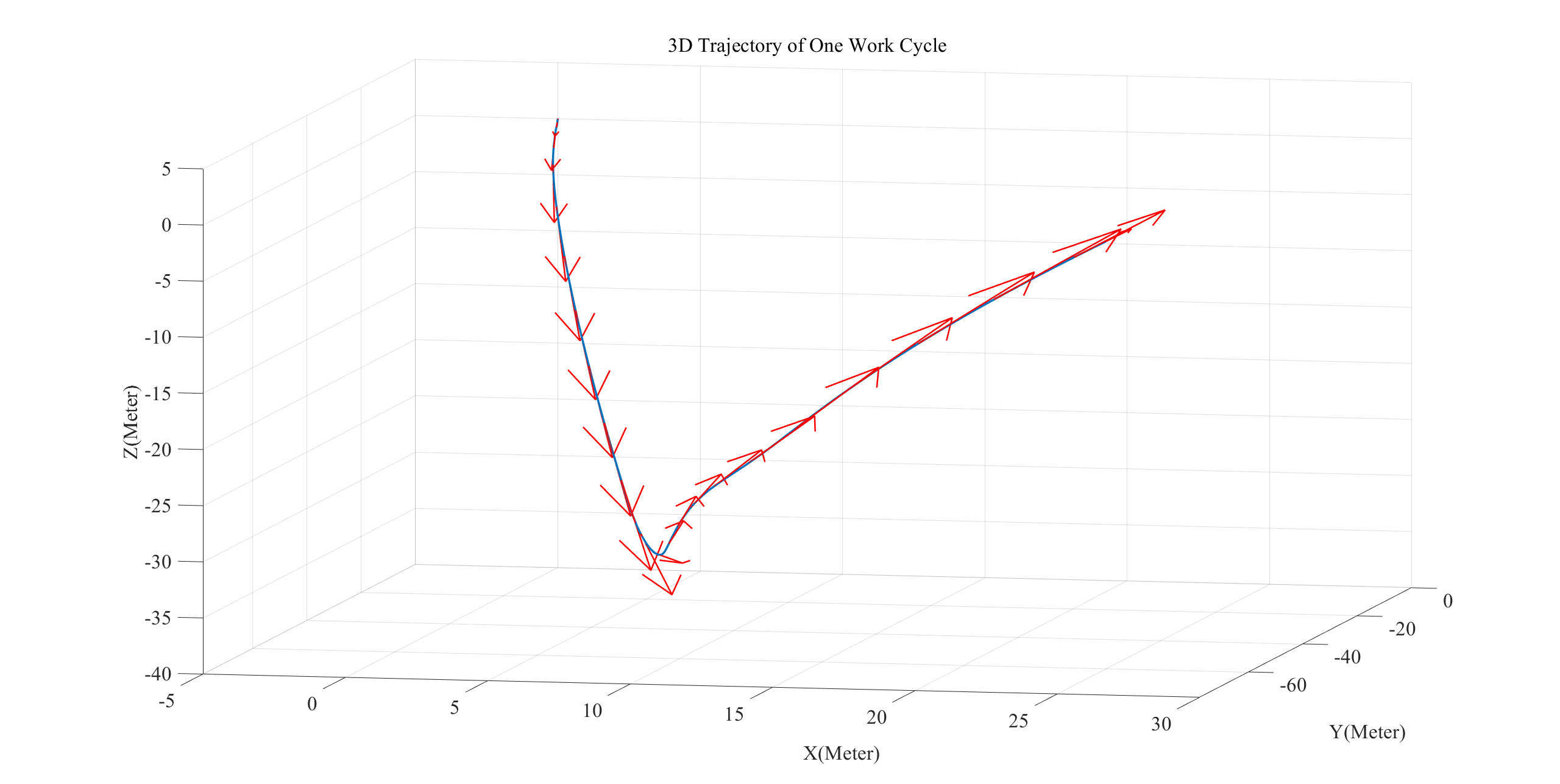}
    \caption{3D Trajectory of One Work Cycle}
    \label{fig:3dTrajectory}
\end{figure}

In the example, the configuration of Petrel II is loaded to the simulator via an URDF file. With the Gazebo 3D rendering engine, the floating, the ascending and the descending of Petrel II are presented in Fig.\ref{fig:floating} to Fig.\ref{fig:ascendingGlider}. A 3D profile of it's one work cycle is depicted in Fig.\ref{fig:3dTrajectory}. In the process, the glider is manipulated by the proposed controller, starting from the surface and dividing into 30 meter depth along the desired descending pitch angle $0.6$ radians, then going up along the desired ascending pitch angle $0.7$ radians. Red arrows refer to direction vectors of velocities of the glider. The coordinates has been transformed into a special NED frame, of which the z-axis is reversed(depth coordinates are negative, instead of being positive), in the sake of clear presentation. The LQR control strategy and the recursive guidance system are implemented on Petrel II as well. 
In the section, the control module would subscribe model states from the Gazebo server directly. The guidance module would receive data from the GPS sensor plugin with the free noise.  

\subsection{Orientation and Velocity Control}

\begin{figure}[H]
    \centering
    \includegraphics[width=0.8\linewidth]{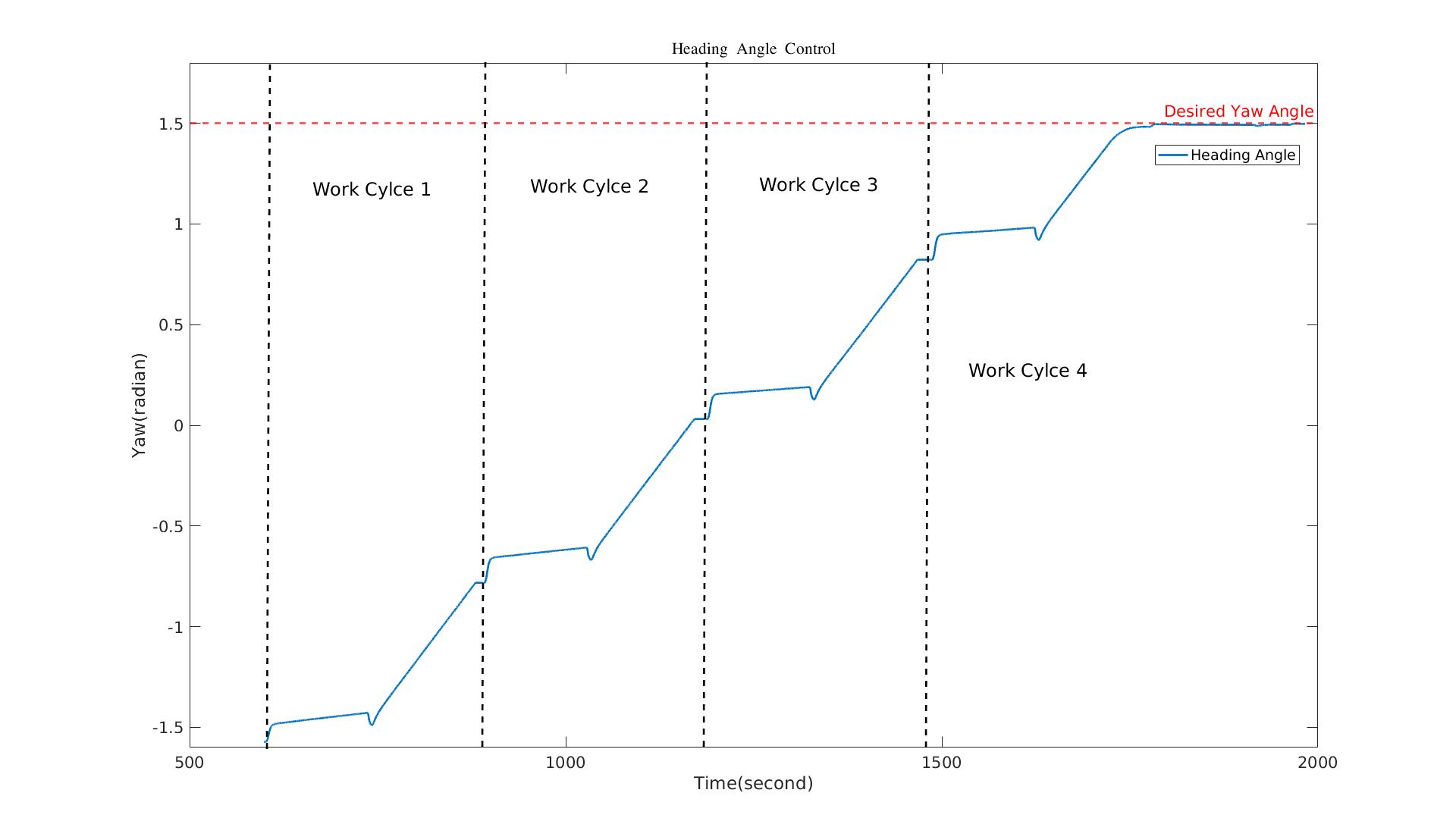}
    \caption{Heading Angle Control}
    \label{fig:yawControl}
\end{figure}

The regulation of the orientation and the forward velocity of is implemented via the LQR control module. As buoyancy-driven gliders lack the good manoeuvrability, the yaw control process is chronic, and lasts for multiple work cycles before reaching the desired angle. As shown in Fig.\ref{fig:yawControl}, the desired yaw angle was $1.5$ radians, the glider might take over 1000 seconds to accomplish the goal from $-1.5$ radians. Within the process, the glider had experienced four work cycles, and each variation of the pitch angle affected the heading angle(or yaw angle), because of the gravity centre changing dramatically.

\begin{figure}[H]
    \centering
    \includegraphics[width=0.8\linewidth]{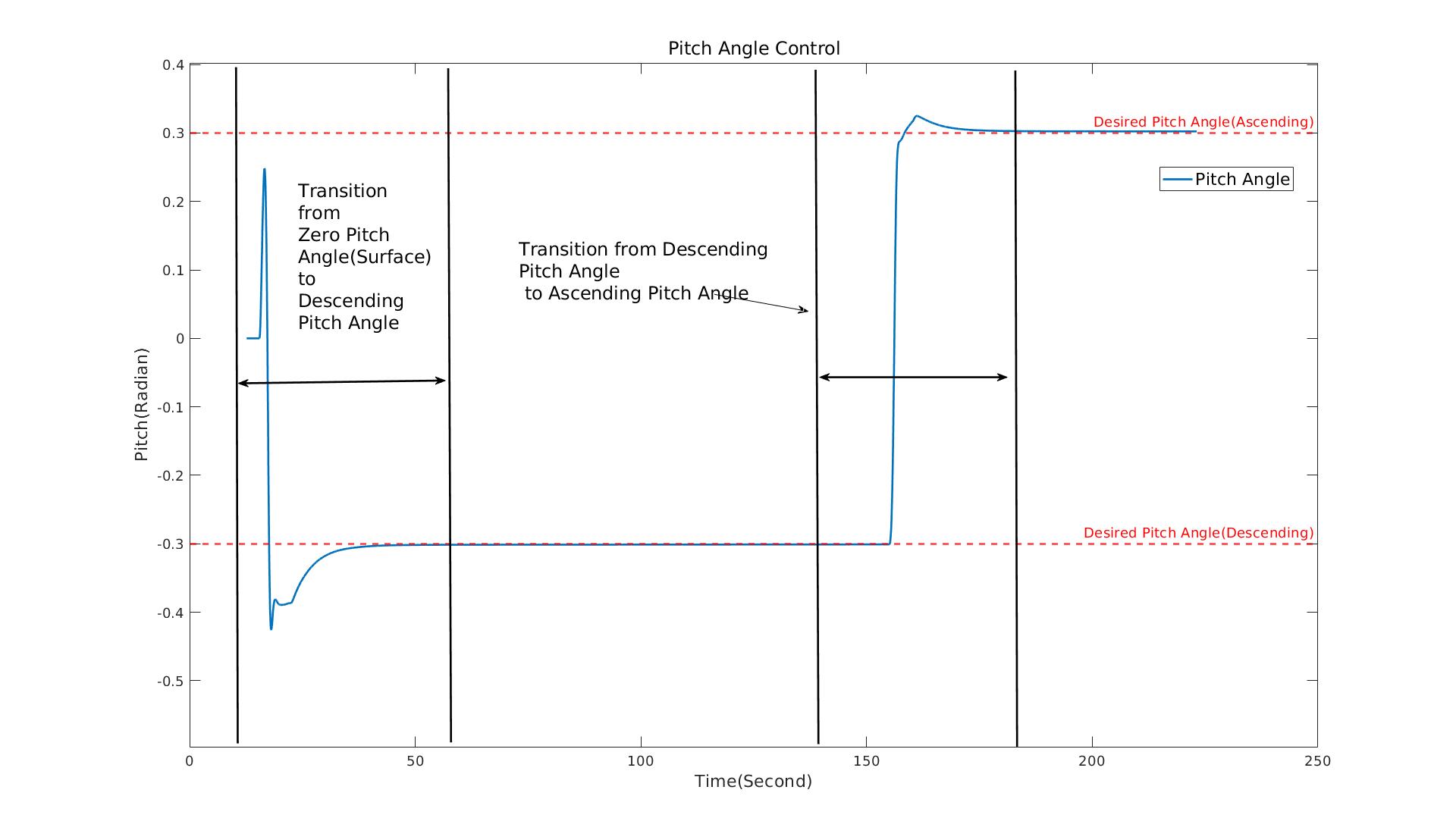}
    \caption{Pitch Angle Control}
    \label{fig:pitchAngleControl}
\end{figure}

The pitch angle control is much more smooth than that of yaw angle, since it is affected directly from the gravity centre which is adjusted through the translational position of the battery package and the ballast mass. Fig.\ref{fig:pitchAngleControl} shows the variation of the pitch angle after the controller was given $-0.3$ desired pitch angle. Due to the changing of the gravity centre, the glider stopping on surface had short-time oscillations about the y-axis of its body-fixed frame and got the desired pitch angle within 30 seconds. When the glider arrived at the desired depth, the pitch angle was regulated to $0.3$ within 20 seconds, to make sure the glider tilt up. Once the glider was submerged, water would give the glider the additional damping. Hence, in the second transition, the oscillation is smaller. 

In the instance, the pitch angle control range is from $-0.6432$ to $0.755$ radians. This is because the restriction of the linear translation of the battery package and the ballast mass, coming from Petrel-II's mechanical layout. 

\begin{figure}[H]
    \centering
    \includegraphics[width=0.8\linewidth]{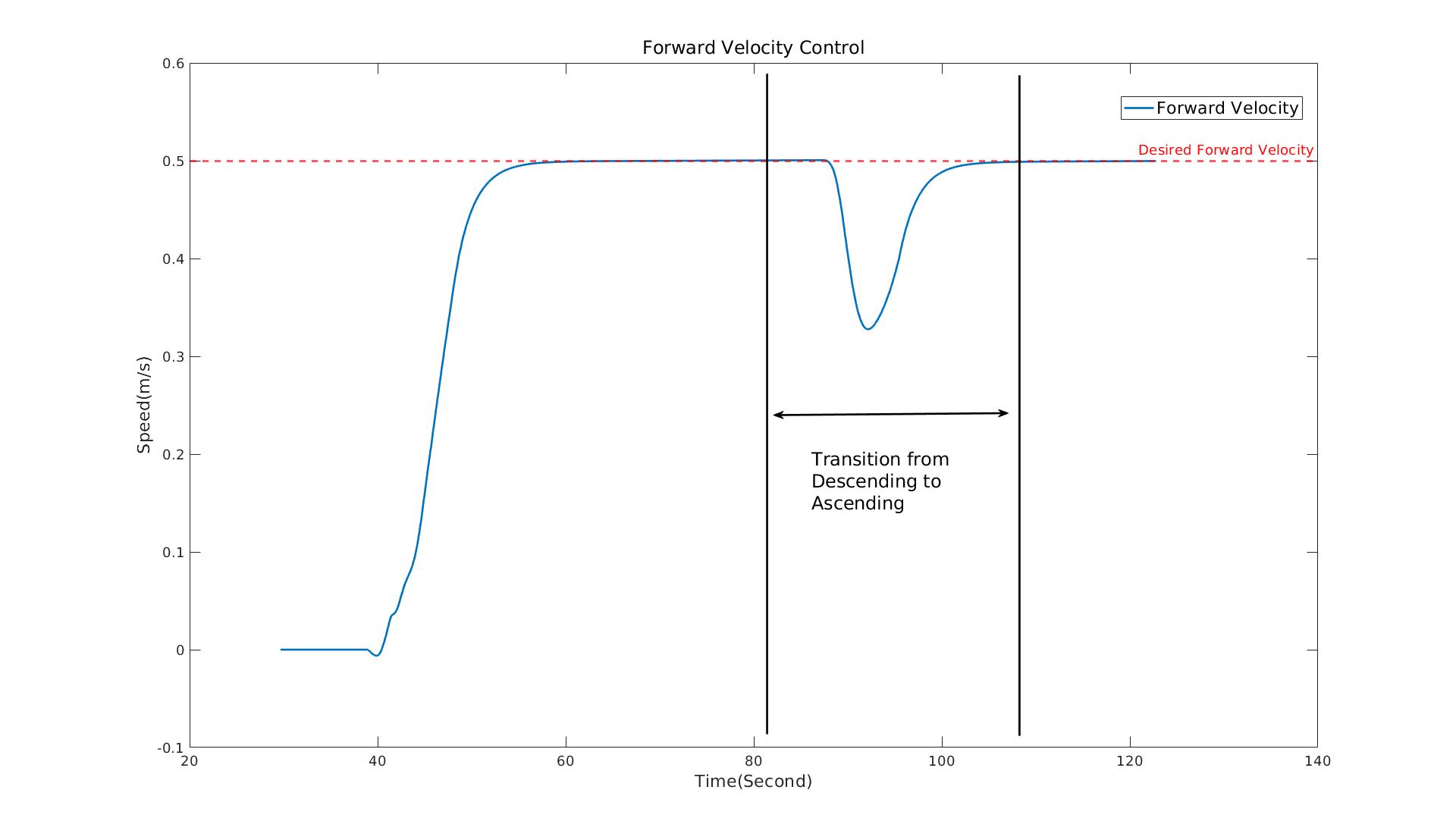}
    \caption{Forward Velocity Control}
    \label{fig:forwardVelocityControl}
\end{figure}

The forward(x-axis on its body frame) speed control is still implemented by the LQR control module. The desired forward speed was set $0.5$ m/s, which was reached within 20 seconds, as shown in Fig.\ref{fig:forwardVelocityControl}. The speed was affected by the transition from the descending to the ascending, for the the orientation and the gravity centre of the gilder was changing significantly during the time. In practical situations, the forward speed control is not essential, particularly when the glider dose not equip with the DVL(Doppler Velocity Log), in order to achieve the low electric power consumption.

\subsection{Waypoints Tracking}

\begin{figure}[H]
    \centering
    \includegraphics[width=0.8\linewidth]{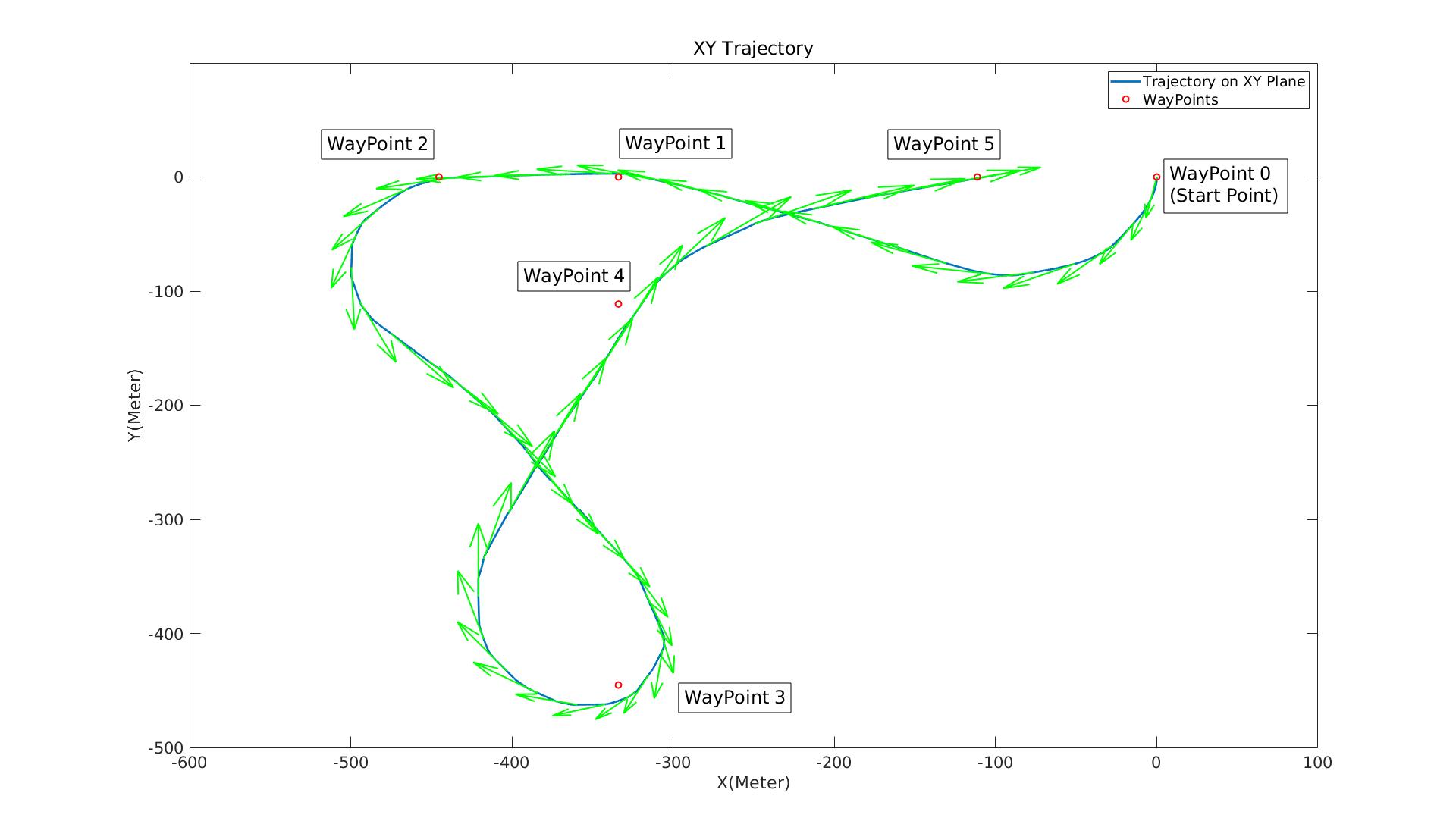}
    \caption{Trajectory on XY-Plane of Waypoints Tracking with $R = 15m$}
    \label{fig:alltask2XY}
\end{figure}

\begin{figure}[H]
    \centering
    \includegraphics[width=0.8\linewidth]{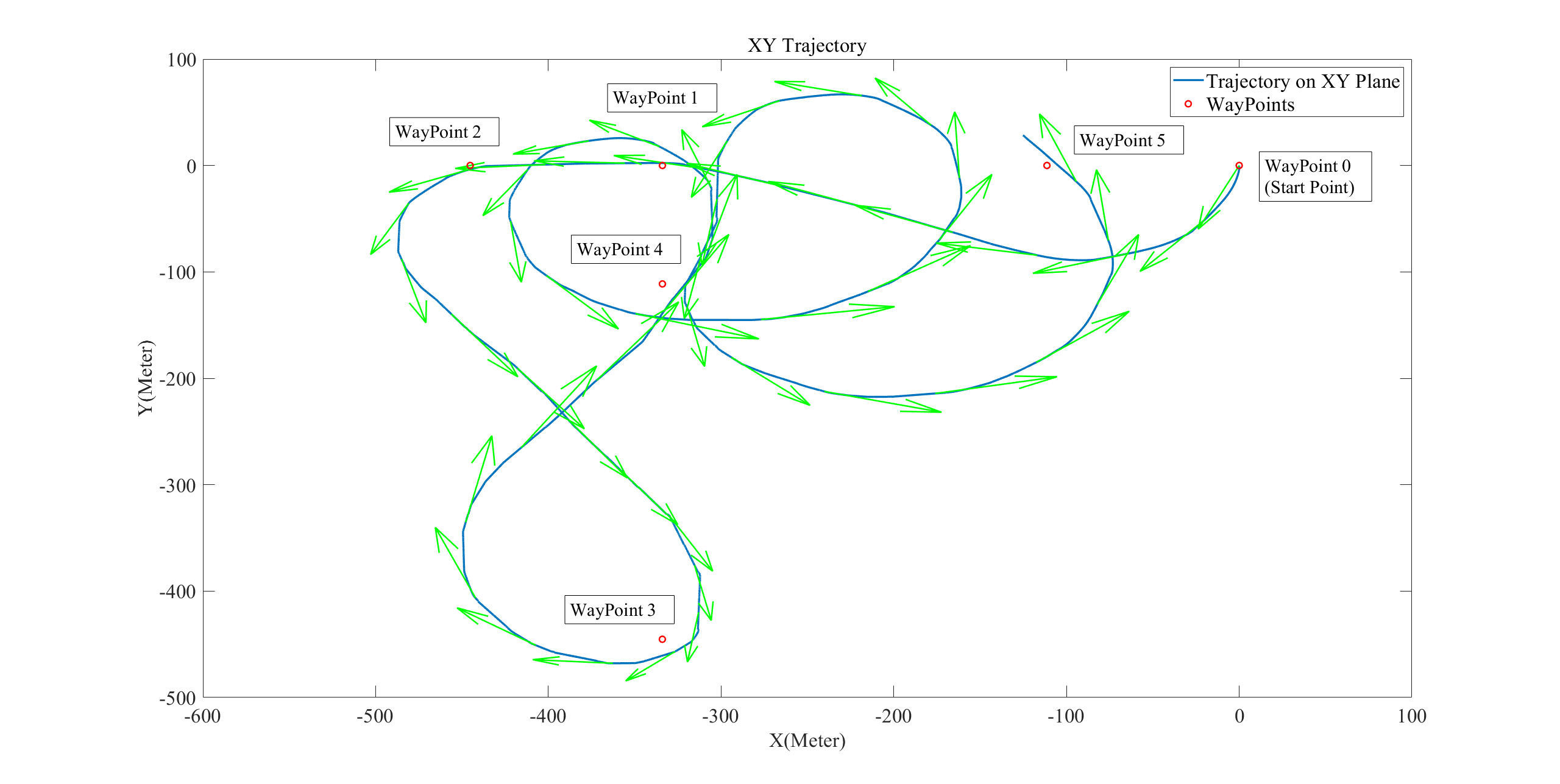}
    \caption{Trajectory on XY-Plane of Waypoints Tracking with $R = 10m$}
    \label{fig:xyTrajectory}
\end{figure}

The waypoints tracking is realised by the recursive guidance system. Due to the low manoeuvrability and the lack of high precision underwater positioning means, the buoyancy-driven glider would update its desired heading angle and pitch angles after each work cycle. There were five waypoints given, as shown in Fig.\ref{fig:alltask2XY} and Fig.\ref{fig:xyTrajectory}, the green arrows indicate the direction of the velocity at that position. The glider was able to go through all waypoints smoothly with the constant circle of acceptance $R = 15m$. However, when the constant circle of acceptance was switched to $R = 10m$, there was a failure, when the glider attempted to get in the adaptive circle of acceptance of the waypoint 4, because of low manoeuvrability. Hence, the glider turned around and made a second attempt to the target. 

\section{Conclusion}

In the paper, a novel simulator for buoyancy-driven gliders is proposed, which brings hydrodynamics and hydrostatics effects into the simulation. This work consisting of the kinetic module, the LQR control module, the recursive guidance module and the tool aiding to design waypoints, will accelerate the software development for low-manoeuvrability buoyancy-driven gliders. We use the Petrel-II glider as an example to present these features. Note that all of those are developed on ROS1 and Gazebo Classic using C++ language, since some sensors' package compatibility issues. In future, the simulator will be rewritten using Gazebo Sim and ROS2 software frameworks.





%





\ifCLASSOPTIONcaptionsoff
  \newpage
\fi



\bibliographystyle{IEEEtran}
\bibliography{bibtex/bib/paper}
\end{document}